\documentclass[11pt]{article}          

\usepackage[margin=1in]{geometry}

\usepackage{graphicx}
\usepackage{times}
\usepackage{amsfonts,amssymb,amsthm,amsmath}
\usepackage{float}
\usepackage{mathtools}
\usepackage{mathrsfs}
\usepackage{bm}
\usepackage{hyperref}
\usepackage[T1]{fontenc}
\usepackage{fancyvrb}
\usepackage{xcolor}
\usepackage{algorithm,url}
\usepackage{epstopdf}

\graphicspath{{.}{figs/}}

\numberwithin{equation}{section}
\xdefinecolor{gray}{rgb}{0.4,0.4,0.4}
\xdefinecolor{blue}{RGB}{58,95,205}

\newcommand{\z}{z}
\newcommand{\el}{\mathcal{E}}
\newcommand{\im}{\mathcal{I}}

\newcommand{\taun}{{\tau}}
\newcommand{\cA}{\mathcal{A}}
\newcommand{\cD}{\mathcal{D}}
\newcommand{\cE}{\mathcal{E}}
\newcommand{\cR}{\mathcal{R}}
\newcommand{\cK}{\mathcal{K}}

\newcommand{\PP}{\mathbb{P}}
\newcommand{\sse}{\sigma_{\text{se}}}
\newcommand{\eps}{\epsilon}
\newcommand{\xv}{\mathbf{x}_v}
\newcommand{\yv}{\mathbf{y}_v}
\newcommand{\bx}{\mathbf{x}_{1:n}}
\newcommand{\by}{\mathbf{y}_{1:n}}
\newcommand{\cby}{\check{\bm{y}}_{1:n}}
\newcommand{\xnew}{x_{n+1}}

\DeclareMathOperator\probit{probit}
\DeclareMathOperator\sgn{sign}
\DeclareMathOperator\Cov{Cov}

\usepackage{natbib}
\setcitestyle{authoryear, open={(},close={)}}

\newtheorem{lemma}{Lemma}[section]
\newtheorem{proposition}{Proposition}[section]
\theoremstyle{remark}
\newtheorem{remark}{Remark}[section]

\begin{document}

\title{Evaluating Gaussian Process Metamodels and Sequential Designs for Noisy Level Set Estimation
}


\author{Xiong Lyu\thanks{Department of Statistics and Applied Probability, University of California UCSB, Santa Barbara, CA 93106-3110, USA 
	\url{lyu@pstat.ucsb.edu}}         \and
        Micka\"el Binois\thanks{rgonne National Laboratory, Mathematics and Computer Science Division,
	9700 South Cass Avenue, Lemont, IL 60439, USA
	\url{mickael.binois@inria.fr}	}  \and
        Michael Ludkovski\thanks{Department of Statistics and Applied Probability, University of California UCSB, Santa Barbara, CA 93106-3110, USA
	\url{ludkovski@pstat.ucsb.edu}}
}




\maketitle

\begin{abstract}
We consider the problem of learning the level set for which a noisy black-box function exceeds a given threshold. To efficiently reconstruct the level set, we investigate Gaussian process (GP) metamodels. Our focus is on strongly stochastic samplers, in particular with heavy-tailed simulation noise and low signal-to-noise ratio. To guard against noise misspecification, we assess the performance of three variants: (i) GPs with Student-$t$ observations; (ii) Student-$t$ processes (TPs); and (iii) classification GPs modeling the sign of the response. In conjunction with these metamodels, we analyze several acquisition functions for guiding the sequential experimental designs, extending existing stepwise uncertainty reduction criteria to the stochastic contour-finding context. This also motivates our development of (approximate) updating formulas to efficiently compute such acquisition functions. Our schemes are benchmarked by using a variety of synthetic experiments in 1--6 dimensions. We also consider an application of level set estimation for determining the optimal exercise policy of Bermudan options in finance.
\end{abstract}

\section{Introduction} \label{introduction}

\subsection{Statement of Problem} \label{statement}
Metamodeling has become widespread for approximating expensive black-box functions that arise in applications ranging from engineering to environmental science and finance \citep{santner2013design}.	 Rather than aiming to capture the precise shape of the function over the entire region, in this article we are interested in estimating the \emph{level set} where the function exceeds some particular threshold. Such problems are common in cases where we need to quantify the reliability of a system or its performance relative to a benchmark. It also arises intrinsically in control frameworks where one wishes to rank the pay-off from several available actions~\citep{hu2015sequential}.

We consider a setup where the latent $f: D \rightarrow \mathbb{R}$ is a continuous function over a $d$-dimensional input space $D \subseteq  \mathbb{R}^d$. The level-set estimation problem consists in classifying every input $x \in D = S \cup N$ according to 	
\begin{align}\label{eq:objective}
S &= \{x \in D: f(x) \geq 0 \}, \qquad N = \{x \in D: f(x)< 0\}.
\end{align}
Without loss of generality the threshold is taken to be zero, so that the level set estimation is equivalent to learning the sign of the response function $f$. For later use we also define the corresponding zero-contour of $f$, namely the partition boundary $\partial S = \partial N = \{x \in D :f(x)=0\}$.

For any $x\in D$, we have access to a simulator $Y(x)$ that generates noisy samples of $f(x)$:
\begin{align}
Y(x) &= f(x)+\epsilon(x), \label{fundamental}
\end{align}
where $\epsilon(x)$ are realizations of independent, mean zero random variables with variance $\taun^2(x)$.

To assess a level-set estimation algorithm, we compare the resulting estimate $\hat{S}$ with the true $S$ in terms of their symmetric difference. Let $\mu$ be a probability measure on the Borel $\sigma$-algebra $\bm{\mathcal{B}}(D)$ (e.g.,~$\mu=\text{Leb}_D$). Then  our loss function is
\begin{align}
L(S,\hat{S}) &= \mu(S\Delta \hat{S}), \label{loss}
\end{align}
where  $S_1 \Delta S_2 := (S_1 \cap S_2^C) \bigcup (S_1^C \cap S_2)$.
Frequently, the inference is carried out by first producing an estimate $\hat{f}$ of the response function; in that case we take $\hat{S}=\{x \in D:\hat{f}(x) \geq 0\}$) and rewrite the loss  as
\begin{align}\label{eq:loss-f}
L(f, \hat{f})   &= \int_{x \in D} \mathbb{I} (\sgn \hat{f}(x) \neq \sgn f(x)) \mu(dx),
\end{align}
where $\mathbb{I}(\cdot)$ is the indicator function.

\subsection{Motivation} \label{motivation}

As a concrete example of level set estimation, consider the problem of evaluating the probability of failure,
determined via the limit state $S$ of a performance function $f(\cdot)$ \citep{picheny2013nonstationary}. The system is safe when $f(x) \le h$, and fails otherwise. In the context where the performance function can be evaluated via deterministic experiments, the estimation of the safe zone (more precisely its volume $\mu(S)$) was carried out in  \cite{bect2012sequential} and \cite{mukhopadhyay2005modeling} employing a Gaussian Process approach with a sequential design. A related example dealing with the probability of failure in a nuclear fissile chain reaction appeared in \cite{chevalier2014fast}.

Another application, which motivated this present investigation, comes from simulation-based algorithms for valuation of Bermudan options \citep{gramacy2015sequential,ludkovski2015kriging}. This problem consists of maximizing the expected reward $h(\tau,X_\tau)$ over all stopping times $\tau \in \{0, \Delta t, 2\Delta t, \ldots, T\}$ bounded by the specified horizon $T$:
\begin{align}
V(t,x) &= \text{sup}_{\tau \geq t, \tau \in \mathcal{S}} \mathbb{E}[h(\tau,X_\tau) | X_t= x], \label{payoff}
\end{align}
where $(X_t)$ is the  underlying asset price at time $t$, typically satisfying a stochastic differential equation and $\Delta t$ is the frequency of exercising. The approach in the so-called Regression Monte Carlo methods \citep{longstaff2001valuing,tsitsiklis2001regression} is to convert the decision of whether to exercise the option $\tau(t,x) = t$ or continue $\tau(t,x) > t$ when $X_t = x$ at intermediate step $t$, into comparing the immediate reward $h(t,x)$ vis \`a vis the reward-to-go $C(t,x)$. In turn this is equivalent to determining the zero level set (known as the continuation region) $S_{t} = \{ x \in D : f(x; t) \ge 0 \}$ of the timing value $f(x; t) := C(t,x) - h(t,x)$. The stopping problem \eqref{payoff} is now solved recursively by backward induction over $t=T-\Delta t,T - 2\Delta t, \ldots$, which allows noisy samples of $f(x; t)$ to be generated by simulating a trajectory  $X^x_{t:T}$ emanating from $x$ and evaluating the respective \emph{pathwise} reward-to-go. Probabilistically, this means that we are interested in \eqref{fundamental} where $f$ corresponds to a \emph{conditional expectation} related to a path-dependent functional of the Markov process $X_{\cdot}$; the loss function \eqref{loss} arises naturally as a metric regarding the quality of the estimated stopping rule in terms of the underlying distribution $\mu(\cdot; t)$ of $X_t$. We refer to \cite{ludkovski2015kriging} for a summary of existing state of the art and the connection to employing a GP metamodel for learning the timing value $T(\cdot; t)$.

\subsection{Design of Experiments for Contour Finding}

Reconstructing $S$ via a metamodel can be divided into two steps: the construction of the response model and the development of methods for efficiently selecting the simulation inputs $x_{1:N}$, known as design of experiments (DoE). Since the level set is intrinsically defined in terms of the unknown $f$, an \emph{adaptive} DoE approach is needed that selects $x_n$'s sequentially.

For the response modeling aspect, GP regression, or kriging, has emerged as the most popular nonparametric approach for both deterministic and stochastic black-box functions \citep{bect2012sequential,gramacy2009adaptive,picheny2013quantile,jalali2016comparison}. GPs have also been widely used for the level-set estimation problem; see
\cite{bryan2008actively,gotovos2013active,hu2015sequential,picheny2010adaptive} and \cite{ranjan2012sequential}.
In a nutshell, at step $n$ the GP paradigm constructs a metamodel $\hat{f}^{(n)}$ that is then used to guide the selection of $x_{n+1}$ and also to construct the estimate $\hat{S}^{(n)}$. To this end, GPs are well suited for sequential design by offering a rich uncertainty quantification aspect that can be (analytically) exploited to construct information-theoretic DoE heuristics. The standard framework is to develop an acquisition function $\im_n(x)$ that quantifies the value of information from taking a new sample at input $x$ conditional on an existing dataset $(x_{1:n}, y_{1:n})$ and then to myopically maximize $\im_n$:
\begin{align}
x_{n+1} = \arg \max_{x \in D } \im_{n}(x). \label{seq}
\end{align}
Early level-set sampling criteria were proposed by~\cite{bichon2008efficient}, \cite{picheny2010adaptive}, and \cite{ranjan2012sequential} based on modifications to the Expected Improvement criterion~\citep{jones1998efficient} for response function optimization. A criterion more targeted to reduce the uncertainty about $S$ itself was first developed by ~\cite{bect2012sequential} using the concept of stepwise uncertainty reduction (SUR). Specifically, the SUR strategy aims to myopically maximize the global learning rate about $S$; see also~\cite{chevalier2014fast} for related computational details.
Recently, further criteria using tools from random set theory were developed in \cite{chevalier2013estimating,azzimonti2015quantifying}. Specifically, those works use the notions of Vorob'ev expectation and Vorob'ev deviation to choose inputs that minimize the posterior expected distance in measure between the level set $S$ and its estimate $\hat{S}$. This  approach is  computationally expensive however, and requires conditional simulations of the posterior Gaussian field. {Other works dealing with more conservative estimates are \cite{bolin2015excursion,azzimonti2016adaptive}.} Clear analysis comparing all these choices in the stochastic setting is currently lacking.

\subsection{Summary of Contributions} \label{summary approach}

{Most} of the cited papers consider only the deterministic setting without any simulation noise. The main goal of this article is to present a comprehensive assessment of GP-based surrogates for stochastic contour-finding. In that sense, our analysis complements the work of~\cite{picheny2013benchmark} and~\cite{jalali2016comparison}, who benchmarked GP metamodels for Bayesian optimization (BO) where the objective is to evaluate $\max_x f(x)$.

While simple versions (with constant or prespecified Gaussian noise) are easily handled, the literature on GP surrogates for complex stochastic simulators remains incomplete. Recently, several works focused on heteroskedastic simulation variance; see the Stochastic Kriging approach of~\cite{ankenman2010stochastic} and the earlier works by two of the  authors~\citep{binois2016practical,binois2017replication}. In the present article we instead target the non-Gaussian aspects, in particular the likely heavy-tailed property. This issue is fundamental to any realistic stochastic simulator where there is no justification for assuming Gaussian-distributed $\epsilon(x)$ (as opposed to the physical experimental setup where $\epsilon$ represents observation noise and is expected to be Gaussian thanks to the central limit theorem). This motivates us to study \emph{alternative GP-based metamodels} for learning $\hat{S}$ that are more robust to non-Gaussian $\epsilon$ in \eqref{fundamental}. In parallel, we investigate which of the contour-finding heuristics outlined above perform best in such setups.

To stay within the sequential design paradigm, we continue to work with a GP-based setup but investigate several modifications that are relevant for learning $\hat{S}$.
\begin{itemize}
	\item To relax the Gaussian noise assumption, we investigate $t$-observation GPs \citep{rasmussen2006gaussian,jylanki2011robust}; the use of the Student-$t$ likelihood nests both the heavy-tailed and Gaussian cases.
	
	\item As another non-Gaussian specification we consider Student-$t$ processes (TPs) \citep{shah2014student,wang2017extended}, as one replacement of GPs, that are also resistant to observation outliers.
	
	\item To target the classification-like objective underlying \eqref{loss}, we consider the use of classification GPs that model the sign of the response $Y(x)$ via a probit logistic model driven by a latent GP $Z(\cdot)$: $\mathbb{P}( Y(x) > 0 | x) = \probit (Z(x))$. Deployment of the logistic regression is expected to ``wash out'' non-Gaussian features in $\epsilon(x)$ beyond its effect on the sign of the observations.
	
	\item In a different vein, to exploit a structure commonly encountered in applications where the level set $S$ is \emph{connected}, we study the performance of \emph{monotone} GP regression/classification metamodels \citep{riihimaki2010gaussian}  that force $f$ (or $Z$) to be monotone in the specified coordinates.
	
\end{itemize}

Our analysis is driven by the primal effect of noise on contour-finding algorithms. This effect was already documented in related studies, such as that of~\cite{jalali2016comparison} who observed the strong impact of $\epsilon(\cdot)$ on performance of BO. Consequently, specialized metamodeling frameworks and acquisition functions are needed that can best handle the stochasticity for the given loss specification.  Thus, the combination of the above tools with the GP framework aims to strike the best balance in carrying out uncertainty quantification  and constructing a robust surrogate that is not too swayed by the simulation noise structure. In the context of GPs, this means accurate inference of the mean response and sampling noise that in turn drive the posterior mean $\hat{f}$ and the posterior GP variance $s(x)^2$. Both of the latter ingredients are needed to blend the exploitation objective to locally learn the contour $\partial S$ and to explore less-sampled regions. These issues drive our choices of the metamodels and also factor in developing the respective acquisition functions $\im_n(x)$; see cf.~Section~\ref{sec:improvementmetrics}. On the latter front we consider four choices (MCU, cSUR, tMSE, ICU), including heuristics that depend only on the posterior standard deviation $s^{(n)}(\cdot)$, as well as those that anticipate information gain from sampling at $x_{n+1}$ via the look-ahead standard deviation $s^{(n+1)}(\cdot)$. Because in the non-Gaussian GPs $s^{(n+1)}$ depends on $Y(x_{n+1})$, we develop tractable approximations $\hat{s}^{(n+1)}$ for that purpose, see Propositions~\ref{updatevarcxt}-\ref{updatevarcxcl}-\ref{updatevarcxtp}.

To recap,
our contributions can be traced along five directions. First, we investigate two ways to handle heavy-tailed simulation noise via a GP with $t$-observations and via TP. As far as we are aware, this is the first application of either tool in sequential design and contour-finding contexts. Second, we present an original use of monotonic GP metamodels for level set estimation. This idea is related to a gray-box approach that aims to exploit known structural properties of $f$ (or $S$) so as to improve on the agnostic black-box strategies. Third, we analyze the performance of classification GP metamodels for contour-finding. This context offers an interesting and novel comparison between regression and classification approaches benchmarked against a shared loss function. Fourth, we develop and implement approximate \emph{look-ahead} formulas for all our metamodels that are used for the evaluation of acquisition functions. To our knowledge, this is the first presentation of such formulas for non-Gaussian GPs, as well as TPs. Fifth, beyond the metamodels themselves, we also provide a detailed comparison among the proposed acquisition functions, identifying the best-performing combinations of $\im(\cdot)$ and metamodel $\hat{f}$ and documenting the complex interplay between design geometry and surrogate architecture.

The rest of the article is organized as follows. Section \ref{sec:model} describes the metamodels we employ. Section \ref{sec:improvementmetrics} develops the sequential designs for the level-set estimation problem, and Section~\ref{sec:update} discusses the look-ahead variance formulas for non-Gaussian GPs. Section~\ref{sec:synthetic} compares the models using synthetic data where ground truth is known. Two case studies from derivative pricing are investigated in Section~\ref{sec:Bermudan}. In Section~\ref{sec:conc} we summarize our conclusions.

\section{Statistical Model}\label{sec:model}

\subsection{Gaussian Process Regression with Gaussian Noise} \label{gauss}


We begin by discussing regression frameworks for contour finding that target learning the latent $f(\cdot)$ based on the loss \eqref{eq:loss-f}. The Gaussian process paradigm treats $f$ as a random function whose posterior distribution is determined from its prior  and the collected samples $\cA_n \equiv \{(x_i,y_i),1 \leq i \leq n\}$. We view $f(\cdot) \sim GP( m(\cdot), K(\cdot,\cdot))$, a priori, as a realization of a Gaussian process completely specified by its mean function $m(x) := \mathbb{E}[f(x)]$ and covariance function $K(x,x') := \mathbb{E}[(f(x)-m(x))(f(x')-m(x'))]$.

In the classical case \citep{rasmussen2006gaussian}, the noise distribution is homoskedastic {Gaussian} $\eps(x) \sim \mathcal{N}(0, \taun^2)$, and the prior mean is zero, {$m(x)=0$}. Given observations  $\by=[y_1, \dots,y_n]^T$ at inputs $\bx=[x_1,\ldots,x_n]^T$, the conditional distribution $f | \cA_n$ is then another Gaussian process, with posterior marginal mean $\hat{f}_{\mathrm{Gsn}}^{(n)}(x_*)$ and covariance $v_{\mathrm{Gsn}}^{(n)}(x_*,x_*')$ given by (throughout we use subscripts to indicate the metamodel type, e.g.,~$Gsn$ for Gaussian noise)
\begin{align}
\hat{f}_{\mathrm{Gsn}}^{(n)}(x_*) &=  k(x_*)[\mathbf{K}+\taun^2\mathbf{I}]^{-1}\by,  \label{mean}\\
v_{\mathrm{Gsn}}^{(n)}(x_*,x_*') &=  K(x_*,x_*')-k(x_*) [\mathbf{K}+\taun^2 \mathbf{I}]^{-1}k(x_*')^T, \label{cov}
\end{align}
with the $1 \times n$ vector $k(x_*)$ and $n \times n$ matrix $\mathbf{K}$ defined by $k(x_*) := K(x_*,\bx) = [K(x_*,x_1),...,K(x_*,x_n)]$, and $\mathbf{K}_{i,j} := K(x_i,x_j)$.

\sloppy The posterior mean $\hat{f}_{\mathrm{Gsn}}^{(n)}(x_*)$ is treated as a point estimate of $f(x_*)$ and the posterior standard deviation $s_{\mathrm{Gsn}}^{(n)}(x_*)^2 = v_{\mathrm{Gsn}}^{(n)}(x_*,x_*)$ as the uncertainty of this estimate. We use $\mathbf{f}$ to denote the random posterior vector $f(\bx) | \cA_n$.

\sloppy \textbf{Model Fitting:} In this article, we model the covariance between the values of $f$ at two inputs $x$ and $x'$ with the squared exponential (SE) function:
\begin{align}
K_{\text{se}}(x,x') := \sse^2\exp\bigg(-\sum_{i=1}^d\frac{(x^i-x'^i)^2} {2\theta_{i}^2}\bigg), \label{covf}
\end{align}
defined in terms of the hyperparameters $\bm{\vartheta}=\{\sse,\theta_1,...,\theta_d, \taun\}$ known as the process variance and length-scales, respectively. Simulation variance $\taun$ is also treated as unknown and part of $\bm{\vartheta}$.
Several common ways exist for estimating $\bm{\vartheta}$. Within a Bayesian approach we integrate against the prior $p(\bm{\vartheta})$ using
\begin{eqnarray}
p(\mathbf{f}|\mathbf{y}_{1:n},\mathbf{x}_{1:n},\bm{\vartheta})&=&\frac{p(\mathbf{y}_{1:n}|\mathbf{x}_{1:n},\mathbf{f})p(\mathbf{f}|\bm{\vartheta})}{p(\mathbf{y}_{1:n}|\mathbf{x}_{1:n},\bm{\vartheta})}, \label{2.1.1}
\end{eqnarray}
where $p(\mathbf{y}_{1:n}|\mathbf{x}_{1:n},\mathbf{f})$ is the likelihood and $p(\mathbf{f}|\bm{\vartheta})$ is the latent function prior. Notice that following the Gaussian noise assumption, the likelihood $p(\mathbf{y}_{1:n}|\mathbf{x}_{1:n},\mathbf{f})$ is Gaussian. With a Gaussian prior $p(\mathbf{f}|\bm{\vartheta})$, the posterior $p(\mathbf{f}|\mathbf{y}_{1:n},\mathbf{x}_{1:n},\bm{\vartheta})$ is tractable and also follows a Gaussian distribution. The normalizing constant in the denominator $p(\mathbf{y}_{1:n}|\mathbf{x}_{1:n},\bm{\vartheta})$ is independent of the latent function and is called the marginal likelihood, given by
\begin{eqnarray}
p(\mathbf{y}_{1:n}|\mathbf{x}_{1:n},\bm{\vartheta})&=&\int p(\mathbf{y}_{1:n}|\mathbf{x}_{1:n},\mathbf{f})p(\mathbf{f}|\bm{\vartheta})d \mathbf{f}. \label{2.1.1.2}
\end{eqnarray}
One may similarly express the posterior over the hyperparameters $\bm{\vartheta}$, where $p(\mathbf{y}_{1:n}|\mathbf{x}_{1:n},\bm{\vartheta})$ plays the role of the likelihood.
To avoid expensive MCMC integration, we use the Maximum Likelihood (ML) estimate $\hat{\bm{\vartheta}}$  which maximizes the likelihood \eqref{2.1.1.2}.
Given the estimated hyperparameters $\hat{\bm{\vartheta}}$, we take the posterior of $f$ as $p(\mathbf{f}|\by,\bx,\hat{\bm{\vartheta}})$.

\subsection{Gaussian Process Regression with Student $t$-Noise} \label{t observation gp}

Taking the noise term $\epsilon(x)$ as Gaussian is widely used since the marginal likelihood is then analytically tractable. In a stochastic simulation setting however, the exact distribution of the outputs relative to their mean is unknown and often is clearly non-Gaussian. A more robust choice is to assume that $\epsilon(x)$ has a Student-$t$ distribution \citep{jylanki2011robust}. In particular, this may work better when the noise is heavy-tailed by making inference more resistant to outliers \citep{o1979outlier}.
In the resulting $t$-GP formulation $\epsilon(x)$ is assumed to be $t$-distributed with variance $\taun^2$ and $\nu > 2$ degrees of freedom (the latter is treated as another hyperparameter). The marginal likelihood of observing $\by$ can be written as
\begin{align}
& p_{t\mathrm{GP}}(\by|\bx, \mathbf{f}) = \prod_{i=1}^n \frac{\Gamma((\nu+1)/2)}{\Gamma(\nu/2)\sqrt{\nu\pi}\sigma_{n}} 
\left(1+\frac{(y_i-f_i)^2}{\nu\sigma_{n}^2}\right)^{-(\nu+1)/2}, \label{liket}
\end{align}
where $\Gamma(\cdot)$ is the incomplete Gamma function.
The likelihood $p_{t\mathrm{GP}}(\mathbf{y}_{1:n}|\mathbf{x}_{1:n},\mathbf{f})$ in~(\ref{2.1.1}) is no longer Gaussian, and integrating (\ref{liket}) against the Gaussian prior $p(f|\bm{\vartheta})$ is intractable; we therefore use the Laplace approximation (LP) method \citep{williams1998bayesian} to calculate the posterior. A second-order Taylor expansion of $\log p_{t\mathrm{GP}}(\mathbf{f}|\bx,\by)$ around its mode, $\tilde{\mathbf{f}}_{t\mathrm{GP}}^{(n)}:=\arg \max_\mathbf{f}p_{t\mathrm{GP}} (\mathbf{f}|\bx,\by)$, gives a Gaussian approximation
\begin{align}\label{lpt}
 p_{t\mathrm{GP}}(\mathbf{f}|\bx,\by) &\approx q_{t\mathrm{GP}}(\mathbf{f}|\bx,\by) 
= \mathcal{N}\left(\tilde{\mathbf{f}}_{t\mathrm{GP}}^{(n)},\mathbf{\Sigma}_{t\mathrm{GP}}^{-1}
\right),
\end{align}
where  $\mathbf{\Sigma}_{t\mathrm{GP}}^{-1}$ is the Hessian of the negative conditional log posterior density at $\tilde{\mathbf{f}}_{t\mathrm{GP}}^{(n)}$:
\begin{align}
\mathbf{\Sigma}_{t\mathrm{GP}} &= -\nabla^2 \log p_{t\mathrm{GP}}(\mathbf{f}|\bx,\by)|_{\mathbf{f}=\tilde{\mathbf{f}}_{t\mathrm{GP}}^{(n)}} 
= \mathbf{K}^{-1}+\mathbf{W}_{t\mathrm{GP}},
\end{align}
and $\mathbf{W}_{t\mathrm{GP}}=-\nabla^2 \log p_{t\mathrm{GP}}(\by|\mathbf{f},\bx)|_{\mathbf{f}=\tilde{\mathbf{f}}_{t\mathrm{GP}}^{(n)}}$ is diagonal, since the likelihood factorizes over observations.

Using \eqref{lpt}, the approximate posterior distribution is also Gaussian $f(x_*)| \cA_n \sim \mathcal{N}( \hat{f}_{t\mathrm{GP}}^{(n)}(x_*), s_{t\mathrm{GP}}^2(x_*))$, defined by its mean $\hat{f}_{\mathrm{t}}^{(n)}(x_*)$ and covariance $v_{t\mathrm{GP}}^{(n)}(x_*,x_*')$:
\begin{align}
\hat{f}_{t\mathrm{GP}}^{(n)}(x_*) &= k(x_*)\mathbf{K}^{-1}\tilde{\mathbf{f}}_{t\mathrm{GP}}^{(n)}, \label{meant} \\
v_\mathrm{t}^{(n)}(x_*,x_*') &= K(x_*,x_*')-k(x_*) [\mathbf{K}+\mathbf{W}_{t\mathrm{GP}}^{-1}] ^{-1}k(x_*'). \label{covt}
\end{align}
Note the similarity to \eqref{mean}--\eqref{cov}:  with Student-$t$ likelihood the mode $\tilde{\mathbf{f}}_{t\mathrm{GP}}^{(n)}$ plays the role of $\by$ and $\bm{W}_{t\mathrm{GP}}^{-1}$ replaces the noise matrix $\tau^2 \mathbf{I}$. Critically, the latter implies that the posterior variance is a function of both designs $\bx$ and observations $\by$.

\subsection{Gaussian Process Classification}
Our target in \eqref{eq:objective} is to learn where the mean response is positive, which is equivalent to classifying each $x \in D$ as belonging either to $S$ or to $N$. Assuming that $\eps(x)$ is symmetric, $\{ x \in S \} = \{ f(x) \ge 0 \} = \{ \mathbb{P}( Y(x) > 0) > 0.5\}$.  This motivates us to consider the alternative of directly modeling the response sign (rather than overall magnitude)  via a classification GP model (Cl-GP)~\citep{williams1998bayesian,rasmussen2006gaussian}.
The idea is to model the probability of a positive observation $Y(x)$ by using a probit logistic regression: $\mathbb{P}(Y(x) > 0 | x) = \Phi( Z(x))$, with $\Phi(\cdot)$ the standard normal cdf.  The latent classifier function is taken as the GP $Z \sim GP( 0, K(\cdot, \cdot) )$. After learning $Z$ we then set $\hat{S} = \{ x \in D: \hat{Z}(x) > 0\}$.

To compute the posterior distribution of $Z$ conditional on $\cA_n$, we use the fact that for an observation $(x_i,y_i)$ and conditional on $z_i = Z(x_i)$ the likelihood of $y_i  > 0$ is $\Phi( \z_i) 1_{\{y_i  \ge 0 \}} + (1-\Phi( \z_i)) 1_{\{y_i < 0\} }$. To simplify notation we use $\check{Y}(x) = \sgn Y(x) \in \{-1, 1\}$ to represent the signed responses driving Cl-GP,  leading to $p_{\mathrm{Cl}}(\cby|\mathbf{\z},\bx)=\prod_{i=1}^n \Phi(\check{y}_i\z_i)$. The posterior of the latent $\mathbf{\z} = Z(x_{1:n})$ is
therefore
\begin{align}
p_{\mathrm{Cl}}(\mathbf{\z}|\bx,\cby) &= \frac{p(\mathbf{\z}|\bx)\prod_{i=1}^n \Phi(\check{y}_i\z_i)}{p(\cby|\bx)}. \label{posteriorz}
\end{align}

Similar to $t$-GP, we use a Laplace approximation for the non-Gaussian $p_{\mathrm{Cl}}(\mathbf{\z}|\bx,\cby)$ in Eq.~(\ref{posteriorz}) (details to be found in Appendix \ref{app:clgp}). The posterior mean for $Z(\cdot)$ at $x_*$ is {\color{blue}then} expressed by using the GP predictive mean equation \eqref{mean} and LP approximation \eqref{cls}:
\begin{align}
\hat{\z}^{(n)}(x_*) &=  k(x_*)\mathbf{K}^{-1}\tilde{\mathbf{\z}}^{(n)}, \label{meanz} \\
v_{\mathrm{Cl}}^{(n)}(x_*,x_*') &=  K(x_*,x_*')-k(x_*)[\mathbf{K}+\mathbf{V}^{-1}]^{-1}k(x_*')^T. \label{covz}
\end{align}
We again see the same algebraic structure, with $\tilde{\mathbf{\z}}^{(n)}$ a stand-in for $\by$ in \eqref{mean} and $\mathbf{V}^{-1}$ a stand-in for $\tau^2 \mathbf{I}$ in \eqref{cov}.
Also note that we may formally link the $Z$ of the Cl-GP metamodel to the GP $f$ used previously via the  posterior probability that $x \in S$:
\begin{align}
\begin{split}
\PP(f(x) \ge 0 | \cA_n) &= \PP(Y(x) > 0|\cA_n)
 = \int_{\mathbb{R}} \Phi(z) p_{Z(x)}(z|\cA_n) dz  \\
&= \int \Phi(z) \phi \bigg(\frac{z - \hat{z}^{(n)}(x)}{{s_{\mathrm{Cl}}^{(n)}(x)}}\bigg) dz
 = \Phi\bigg(\frac{\hat{\z}^{(n)}(x)}{\sqrt{1+s_{\mathrm{Cl}}^{(n)}(x)^2}}\bigg).
\end{split}
\end{align}

{
	\subsection{Student-$t$ Process Regression with Student-$t$ Noise}
	Instead of just adding Student-$t$ likelihood to the observations, \cite{shah2014student} proposed $t$-processes (TPs) as an alternative to GPs, deriving closed-form expressions for the marginal likelihood and posterior distribution of the $t$-process by imposing an inverse Wishart process prior over the covariance matrix of a GP model. They found the $t$-process to be more robust to model misspecification and to be particularly promising for BO. Moreover, \cite{shah2014student} showed that TPs retain most of the appealing properties of GPs, including analytical expressions, with increased flexibility.

	As noticed for example in \cite{rasmussen2006gaussian}, dealing with noisy observations is less straightforward with TPs, since the sum of two independent Student-$t$ distributions has no closed form. Still, this drawback can be circumvented by incorporating the noise directly in the kernel. The corresponding data-generating mechanism is  taken to be multivariate-$t$ $\mathbf{y}_{1:n} \sim \mathcal{T} \left(\nu, m(\bx), \mathbf{K}+\taun^2 \mathbf{I} \right)$, where the degrees of freedom are $\nu \in (2,\infty)$. The posterior predictive distribution is then $f(x_*)|\mathcal{A}_n \sim \mathcal{T}\left(\nu + n, \hat{f}^{(n)}_{\mathrm{TP}}(x_*), v^{(n)}_{\mathrm{TP}}(x_*, x_*) \right)$, where \citep{shah2014student}
	\begin{align}
	\hat{f}_{\mathrm{TP}}^{(n)}(x_*) =&  k(x_*)[\mathbf{K}+\taun^2\mathbf{I}]^{-1}\by,  \label{TPmean}\\
	v_{\mathrm{TP}}^{(n)}(x_*,x'_*) =& \frac{\nu + \beta^{(n)} -2}{\nu + n - 2} \left\{ K(x_*,x_*')  
	 - k(x_*) [\mathbf{K}+\taun^2 \mathbf{I}]^{-1}k(x_*')^T \right\}, \label{TPcov}
	\end{align}
	with $$\beta^{(n)} := \by^\top [\mathbf{K} + \taun^2\mathbf{I}]^{-1} \by.$$
	
	
	Comparing with the regular GPs, we have the same posterior mean $\hat{f}_{\mathrm{TP}}^{(n)}(x_*) = \hat{f}_{\mathrm{Gsn}}^{(n)}(x_*)$, but the posterior covariance now depends on observations $\by$ and is inflated: $v_{\mathrm{TP}}^{(n)}(x_*,x'_*) = \frac{\nu + \beta^{(n)} -2}{\nu + n - 2} v_{\mathrm{Gsn}}^{(n)}(x_*,x'_*)$. Moreover, the latent function $f$ and the noise are uncorrelated but not independent. As noticed in \cite{shah2014student}, assuming the same hyperparameters, as $n$ goes to infinity, the above predictive distribution becomes Gaussian.
	
	Inference of TPs can be performed similarly as for a GP, for instance based on the marginal likelihood:
	\begin{equation}
	 p_{\mathrm{TP}}(\by | \bx, \bm{\vartheta}) = \frac{\Gamma(\frac{\nu + n}{2})}{((\nu-2)\pi)^{\frac{n}{2}} \Gamma(\frac{\nu}{2})} |\mathbf{K}|^{-1/2} 
\left( 1 + \frac{\by^\top \mathbf{K}^{-1} \by }{\nu - 2} \right)^{-\frac{\nu + n}{2}}.
	\end{equation}
}
One issue is estimation of $\nu$, which plays a central role in the TP predictions. We find that restricting $\nu$ to be small is important in order to avoid degenerating to the plain Gaussian GP setup.

\subsection{Metamodel Performance for Level Set Inference} \label{statistics}
To evaluate the performance of different metamodels, we consider several metrics.
The first statistic is the error rate $\mathcal{ER}$ based on the loss function $L$ defined in Eq.~(\ref{loss}), measuring the distance between the level set $S$ and its estimate $\hat{S}$:
\begin{align}
\mathcal{ER} &:= \mu(S\Delta \hat{S}) 
= \int_{x \in D} \mathbb{I} \left[\sgn f(x)\neq \sgn \hat{f}(x)\right] \mu(dx).
\label{erc}
\end{align}
For Cl-GP, we replace $f(x)$ with $z(x)$ in the above, namely, use~$\mu(S \Delta \hat{S}) = \mu\{ x \in D: \hat{z}(x) < 0 < z(x) \cup \hat{z}(x) > 0 > z(x) \}$. A related statistic is the bias $\mathcal{B}$, which is based on the \emph{signed} ($\mu$-weighted) difference between $S$ and $\hat{S}$:
\begin{align}
\mathcal{B}  =& \mu(S\backslash \hat{S}) - \mu(\hat{S}\backslash S) 
= \int_{x \in D} \! \left\{\mathbb{I}[\hat{f}(x)<0 < f(x)]-\mathbb{I}[\hat{f}(x)>0 > f(x)] \right\} \mu(dx).
\label{bci}
\end{align}

The error rate $\mathcal{ER}$ and bias $\mathcal{B}$ evaluate the accuracy of the point estimate $\hat{S}$ when the ground truth is known. In a realistic case study when the latter is unavailable, we replace $\cR$ by its empirical counterpart, based on quantifying the uncertainty in $\hat{S}$ through the associated uncertainty of $\hat{f}$.
Following \cite{azzimonti2015quantifying}, we define the empirical error $\cE$ as the expected distance in measure between the random set $S | \cA$ and its estimate $\hat{S}$:
\begin{align}
\cE := \mathbb{E} \left[\mu(S \Delta \hat{S}) | \, \cA \right]
= \int_{x \in D} \bar{E}(x) \mu(dx),  \label{eec}
\end{align}
with $\bar{E}(x)$ calculated by using 
(\ref{mean}) and (\ref{cov}): 
\begin{align}
\nonumber \bar{E}(x) &:= 
\mathbb{E} \left[\mathbb{I}[\sgn f(x) \neq \sgn \hat{f}(x)] | \cA \right] 
\\
\nonumber &= \int_{\mathbb{R}} \mathbb{I}[\sgn f(x) \neq \sgn \hat{f}(x)]p(f(x)|\cA) df(x) = \Phi\bigg(\frac{-|\hat{f}(x)|}{s(x)}\bigg).\label{empiricalerrorc}
\end{align}
The local empirical error $\bar{E}(x)$ is the posterior probability of wrongly classifying $x$ conditional on the training dataset $\cA$.
It is intrinsically tied to the point estimate $\hat{f}(x)$ and the associated posterior variance $s(x)^2$ through the Gaussian uncertainty quantification. For the TPs, the predictive distribution is Student-$t$, so that the Gaussian cdf $\Phi$ is replaced with the respective survival function.

\textbf{Uncertainty Quantification:}
To quantify the  overall uncertainty about $S$ (rather than local uncertainty about $f(x)$), a natural criterion is the \emph{volume}
of the credible band $CI_{\partial S}$ that captures inputs $x$ whose sign classification remains ambiguous given $\cA$. A simple definition at a credibility level $\alpha$ (e.g.,~$\alpha = 0.05$) would be
\begin{equation}
CI_{\partial S}^{(n)} = \left\{ {x} \in D: \left(\hat{f}^{(n)}({x}) + z_{1-\frac{\alpha}{2}}s^{(n)}({x})\right) 
\left(\hat{f}^{(n)}({x}) - z_{1-\frac{\alpha}{2}}s^{(n)}({x})\right)<0 \right\}, \label{cis}
\end{equation}
where $z_{1-\frac{\alpha}{2}}$ is the appropriate Gaussian/Student-$t$ $\alpha$-quantile. Thus \eqref{cis} evaluates the region where the sign of $f$ is nonconstant over the posterior $\alpha$-CI of $f$. Heuristically however, $CI_{\partial S} \simeq \{x \in D : \bar{E}(x) > {\alpha} \}$ is effectively equivalent to empirical error $\bar{E}(x)$ exceeding $\alpha$, so that the volume of $CI_{\partial S}$ is roughly proportional to the integrated empirical error  $\mathcal{E}$.

In a more sophisticated approach based on random set theory, \cite{chevalier2013estimating} used the Vorob'ev deviation to define the uncertainty measure $V_\alpha(\hat{S})$:
\begin{align}
\nonumber V_\alpha(\hat{S}) :=& \mathbb{E} \left[\mu(\hat{S}^\alpha \Delta S)  | \ \cA \right] \\
\nonumber =& \int_{x \in \hat{S}^\alpha} \mathbb{P}(x \notin S | \cA) \mu(dx) 
+ \int_{x \in (\hat{S}^\alpha)^C} \mathbb{P}(x \in S | \cA) \mu(dx) \\
 =& \int_{x \in \hat{S}^\alpha} \left(1-p_V(x) \right) \mu(dx) + \int_{x \in (\hat{S}^\alpha)^C} p_V(x) \mu(dx), \label{vorob}
\end{align}
where $$
\hat{S}^\alpha := \left\{x \in D: \hat{f}(x)- z_{1-\frac{\alpha}{2}} s(x) \geq 0\right\}$$ and $$p_V(x) = \PP(x \in S | \cA) = \Phi \bigg(\frac{\hat{f}(x)}{s(x)} \bigg).
$$
An $\alpha$ satisfying the unbiasedness condition $$\int_{x \in D} p_V(x) \mu(dx) = \mathbb{E}[\mu(S) | \cA]=\mu(\hat{S}^\alpha)$$ is referred to as the \emph{Vorob'ev threshold} and can be determined through dichotomy \citep{chevalier2013estimating}. If the {Vorob'ev threshold} is picked to be zero, then the Vorob'ev deviation is reduced to the empirical error $\el$.
Because of the computational overhead of working with \eqref{vorob}, we restrict attention to the credible bands defined through $\hat{S}^\alpha$, which correspond to local uncertainty about $f$ (or $Z$) as in \eqref{cis}.

\section{Sequential Design}\label{sec:improvementmetrics}

We estimate the level set $S$ in a sequential design setting that assumes that $f$ is expensive to evaluate, for example because of the complexity of the underlying stochastic simulator. Therefore efficient selection of the inputs $\mathbf{x}_{1:n}$ is important. In sequential design, at each step the next sampling location $x_{n+1}$ is  selected given all previous measurements. The Bayesian approach to sequential design is based on greedily optimizing an acquisition function as in \eqref{seq}. These strategies got popularized thanks to the success of the expected improvement (EI) criterion and the associated efficient global optimization (EGO) algorithm~\citep{jones1998efficient}.
The basic loop for sequential design is as following:

\begin{itemize}
	\item Initialize $\mathcal{A}_{n_0} = \{(x_i,y_i),1 \leq i \leq n_0\}.$
	\item Loop for $n=n_0 {+1},\dots$ $N$.
	\begin{itemize}
		\item Choose the next input $x_{n+1} = \arg\max_{x\in \mathcal{M}} \im_{n}(x)$, and sample $y_{n+1}=Y(x_{n+1})$.
		\item Augment $\mathcal{A}_{n+1}=\mathcal{A}_n \bigcup \{(x_{n+1},y_{n+1})\}.$
		\item Update $\hat{S}^{(n+1)}$ with $\mathcal{A}_{n+1}.$
	\end{itemize}
\end{itemize}

We now propose several metrics for the acquisition function $\im_n(x)$ in Eq.~(\ref{seq}). The key plan is to target regions close to the boundary $\partial \hat{S}$. A second strategy is to use the look-ahead posterior standard deviation $s^{(n+1)}$ conditional on sampling at $x$, in order to assess the corresponding \emph{information gain}. This links the constructed design to the metamodel for $f$, since different surrogate architectures quantify uncertainty differently.

The first metric, dubbed Maximum Contour Uncertainty (MCU), stems from the Upper Confidence Bound (UCB) strategies proposed by~\cite{srinivas2012information} for Bayesian optimization. The idea of UCB is to express the exploitation-exploration trade-off through the posterior mean $\hat{f}(x)$ and standard deviation $s(x)$.  Following the spirit of UCB, MCU blends the minimization of $|\hat{f}^{(n)}(x)|$ (exploitation) with maximization of the posterior uncertainty $s^{(n)}(x)$ (exploration):
	\begin{align}
	\im_n^{\text{MCU}}(x) &:= -|\hat{f}^{(n)}(x)| + \gamma^{(n)} s^{(n)}(x), \label{ucb}
	\end{align}
	where $\gamma^{(n)}$ is a step-dependent sequence of weights. Thus,  MCU targets inputs with high uncertainty (large $s^{(n)}(x)$) and close to the boundary $\partial \hat{S}$ (small $|\hat{f}^{(n)}|$ ). Small $\gamma^{(n)}$ leads to aggressive sampling concentrated along the estimated $\partial \hat{S}$; large $\gamma^{(n)}$ leads to space-filling sampling that effectively minimizes the integrated mean-squared error. Thus, the choice of $\gamma$'s is critical for the performance; in particular $\gamma^{(n)}$ should be increasing to avoid being trapped in local minima of $|\hat{f}^{(n)}(x)|$. In the original application to BO \citep{srinivas2012information} it is proved that with $\gamma^{(n)} = ({2 \log \big (\frac{|D| \pi^2 n^2}{6\delta}\big)})^{1/2}$, the regret (a metric measuring the distance between estimated optima and the trueth in BO) of the estimate is guaranteed to converge. Further recipes for \eqref{ucb} for level set estimation were proposed in~\cite{gotovos2013active} and~\cite{bogunovic2016truncated}; both papers mention that the above recommendation is too conservative and tends to over-explore. A constant choice of $\gamma^{(n)} = 1.96$ corresponds to the Straddle scheme in~\cite{bryan2006active} and leads to $\im_n(x) \ge 0 \Leftrightarrow x \in$ (95\% CI band of $\partial S$). Similarly, \cite{gotovos2013active} employed $\gamma^{(n)} = 3$ and \cite{bogunovic2016truncated} suggested to use $\gamma^{(n)} = \sqrt{\log (|D| n^2)}$. Based on our experiments (see Appendix \ref{app:mcu}), we find that a constant value of $\gamma^{(n)}$ may be problematic and recommend to adapt $\gamma^{(n)}$ to the relative ratio between $f(x)$ (for steeper response surfaces $\gamma$ should be larger) and $s(x)$ ($\gamma$ needs to rise as posterior uncertainty decreases). One recipe is to use $\gamma^{(n)} = IQR(\hat{f}^{(n)}) \backslash 3 Ave(s^{(n)})$ ($Ave(s^{(n)})$ denotes the average of standard deviation and IQR the inter-quantile range) which keeps both terms in \eqref{ucb} approximately comparable as $n$ changes.


\begin{remark}\label{app:mee}
		The local empirical error $\bar{E}(x)$ as defined in Eq.~(\ref{empiricalerrorc}) could be directly used as an acquisition function, i.e.,
		\begin{align}
		\im_n^{\text{MEE}}(x) \equiv \bar{E}(x) = 
		\Phi\bigg(-\frac{|\hat{f}^{(n)}(x)|}{s^{(n)}(x)}\bigg). \label{criterionmee-disc}
		\end{align}
		This Maximal Empirical Error (MEE) acquisition function measures the local probability of misclassification and is similar to the sequential criteria in \cite{bect2012sequential,echard2010kriging,ranjan2012sequential,bichon2008efficient}, all based on the idea of sampling at $x$ where the event $\{f(x) \geq 0\} | \cA_n$ is most uncertain. However, \eqref{criterionmee-disc} is not suitable for our purposes since it is maximized across the entire $\partial \hat{S}$ (namely $\im_n^{\text{MEE}}(x) =0.5$ for any $x$ where $\hat{f}^{(n)}(x) = 0$), so does not possess a unique maximizer as soon as $\partial \hat{S}$ is non-trivial. One potential solution could be to maximize \eqref{criterionmee-disc} over a finite candidate set, which however requires significant fine-tuning.
\end{remark}

Our second strategy focuses on quickly \emph{reducing} $\bar{E}$ by comparing the current $\bar{E}(x)$ given $\cA_n$ and the expected $\bar{E}(x)$ conditional on the one-step-ahead sample, $\cA_n \cup \{ x_{n+1}, y_{n+1} \}$. This is achieved by integrating out the effect of $Y(x_{n+1})$ on $\bar{E}(x_{n+1})$:
\begin{align}\label{criterionmeesur-1}
\begin{split}
\im_n^{\text{cSUR}}(x) =& \im_n^{\text{MEE}}(x) - \mathbb{E}_{Y(x)} \left[ \im_{n+1}^{\text{MEE}}(x) \right] \\
=&
\Phi\bigg(-\frac{|\hat{f}^{(n)}(x)|}{s^{(n)}(x)}\bigg)
-\mathbb{E}_{Y(x)}\bigg[\Phi\bigg(-\frac{|\hat{f}^{(n+1)}(x)|}{s^{(n+1)}(x)}\bigg)\bigg].
\end{split}
\end{align}
The name cSUR is because \eqref{criterionmeesur-1} is directly related to the SUR strategy \citep{bect2012sequential}, modified to target \emph{c}ontour-finding. Crucially, $\im^{cSUR}$ ties the selection of $x_{n+1}$ to the look-ahead mean $\hat{f}^{(n+1)}(x_{n+1})$ and look-ahead standard deviation $s^{(n+1)}(x_{n+1})$ that appear on the right-hand side of \eqref{criterionmeesur-1}.
To compute the integral over $Y(x)$, we replace $\hat{f}^{(n+1)}(x)$ with its average $\hat{f}^{(n)}(x)=\mathbb{E}_n[f(x)]=\mathbb{E}_n[\mathbb{E}_{n+1}[f(x)]]=\mathbb{E}_n[ \hat{f}^{(n+1)}(x)]$. Similarly, we plug in the approximate one-step-ahead standard deviation $\hat{s}^{(n+1)}$ discussed in Section \ref{sec:update} (especially Equations (\ref{updategv}), (\ref{estvt}), and (\ref{estvc})) for $s^{(n+1)}(x)$:
\begin{align}\label{criterionmeesur}
\hat{\im}_n^{\text{cSUR}}(x)  =& \Phi\bigg(-\frac{|\hat{f}^{(n)}(x)|}{s^{(n)}(x)}\bigg) -\Phi\bigg(-\frac{|\hat{f}^{(n)}(x)|}{\hat{s}^{(n+1)}(x)|_{x_{n+1} = x} }\bigg).
\end{align}
Note that if $x$ is such that $\hat{f}^{(n)}(x) = 0$ then both terms above are $1/2$ and $\im_n^{\text{cSUR}}(x) = 0$. Thus, the cSUR criterion will not place samples \emph{directly} on $\partial \hat{S}$, but will aim to bracket the zero-contour.

In \eqref{criterionmeesur} cSUR only measures the \emph{local} improvement in $\bar{E}(x_{n+1})$ at the sampling location $x_{n+1}$ and consequently might be overly aggressive in targeting $\partial \hat{S}$. This motivates us to target the \emph{global} reduction in the uncertainty of $\hat{S}$, so as to take into account the spatial structure of $D$. The resulting Integrated Contour Uncertainty (ICU) is linked to the already defined empirical error $\cE$ from Section \ref{statistics}:
\begin{align}
\begin{split}
& \im_n^{\text{ICU}}(x) := \cE^{(n)} - \mathbb{E}_{Y(x)}[\cE^{(n+1)}|x_{n+1}=x] \\
& \; = \cE^{(n)} -
  \mathbb{E}_{Y(x)} \bigg[\int_{u \in D} \!\Phi\bigg(\frac{-|\hat{f}^{(n+1)}(u)|}{s^{(n+1)}(u)|_{x_{n+1}=x}}\bigg)\mu(du)\bigg].
\end{split}
\end{align}
We apply the same approximation as for cSUR to simplify the expectation over $Y(x)$ and replace the integral over $D$ with a sum over a finite subset $\cD$ of size $M$:
\begin{align}\label{criterioneee}
\hat{\im}_n^{\text{ICU}}(x) & = -\sum_{x_m \in \cD} \Phi\bigg(\frac{-|\hat{f}^{(n)}(x_m)|}{\hat{s}^{(n+1)}(x_m)|_{x_{n+1} = x}}\bigg) \mu(x_m).
\end{align}
Then $\im^{ICU}(x)$ can be viewed as measuring the overall information gain about $S$ from sampling at $x$. The motivation behind ICU is to myopically minimize the expected one-step-ahead empirical error $\mathcal{E}$, which would correspond to 1-step Bayes-optimal design.

As a last alternative, we utilize the targeted mean squared error (tMSE) criterion, a localized form of targeted IMSE criterion in~\cite{picheny2010adaptive}:
\begin{align}
\im_n^{\text{tMSE}}(x) &:= s^{(n)}(x)^2 \cdot W_n^{\text{tMSE}}(x),  \label{criteriontmse}
\end{align}
where
\begin{align}
W_n^{\text{tMSE}}(x) &:= \frac{1}{\sqrt{2 \pi} s^{(n)}(x)} \exp \bigg(-\frac{\hat{f}_n(x)^2}{2s^{(n)}(x)^2}\bigg).
\label{eq:w-tmse}
\end{align}
	
The tMSE criterion upweighs regions close to the zero contour through the weight function $W_n^{\text{tMSE}}(x)$ which measures the distance of $x$ to $\partial \hat{S}^{(n)}$ using the Gaussian posterior density $\mathcal{N}(\hat{f}^{(n)}, s^{(n)}(x)^2)$. Like MCU, tMSE is based only on the posterior at step $n$ and does not integrate over future $Y(x)$'s.

\begin{remark}
	In~\cite{picheny2010adaptive} an additional parameter $\sigma_\epsilon$ was added to the definition of $W_n^{\text{tMSE}}(x)$ by replacing $s^{(n)}(x)$ everywhere with $\sqrt{ s^{(n)}(x)^2 + \sigma_\epsilon^2}$. Larger $\sigma_\epsilon$ yields more space-filling as $W_n^{\text{tMSE}}(x)$ becomes flatter. Since~\cite{picheny2010adaptive} dealt with deterministic experiments, $\sigma_\epsilon$ was necessary to ensure that $W_n^{\text{tMSE}}(x)$ is well defined at existing $x_{1:n}$ and the recommendation was for $\sigma_\epsilon$ to be 5\% of the range of $f$. In our case $s^{(n)}(x)$ is intrinsically bounded away from zero and \eqref{eq:w-tmse} works well as is. Additional experiments (available upon request) indicate that the performance of \eqref{criteriontmse} is not sensitive to $\sigma_\epsilon$, so to minimize the number of tuning parameters we stick to $\sigma_\epsilon=0$ in \eqref{eq:w-tmse}.
\end{remark}

In the TP case, for MCU, cSUR, and ICU, we replace the standard normal cdf $\Phi(\cdot)$ appearing in the formulas by its Student-$t$ counterpart (with the estimated degrees of freedom $\nu_n$). For tMSE, to maintain tractability, we keep the same expression \eqref{eq:w-tmse} for the weights $W^{\text{tMSE}}$.

\begin{figure}[ht]
	\centering
	\includegraphics[height=2.4in]{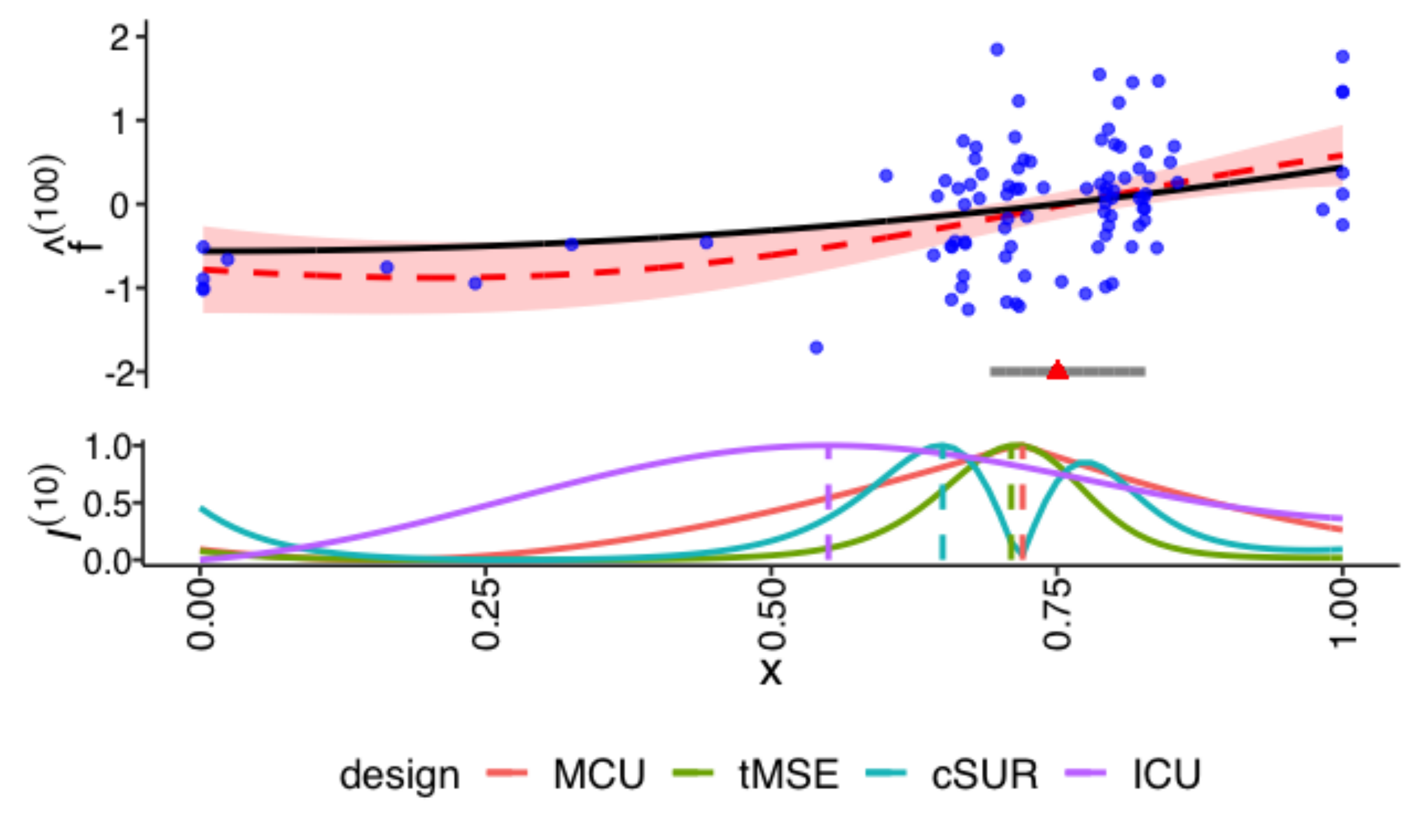}
	\caption{Comparison of acquisition functions. \emph{Upper} panel: true function $f= (x+0.75)(x-0.75)$ (black solid line), the posterior mean $\hat{f}(\cdot)$ (dashed line) and 95\% $CI_f$ (shaded area) based on observed samples $(\mathbf{x}_{1:100},\mathbf{y}_{1:100})$ (blue dots). Along the x-axis we also show the credible interval of the partition boundary $CI_{\partial S}$ (grey solid line) relative the true zero level set $S=[0,0.75]$ (red triangle). \emph{Lower} panel: acquisition functions $\im_n(\cdot)$ for MCU, cSUR, ICU, and tMSE criteria, with vertical lines marking the respective maxima $\arg\max_x\im_n(x)$.}
	\label{acquisition}
\end{figure}

\subsection*{Illustration}

 For instructive purposes, we consider a one-dimensional case where we use the Gaussian observation GP to learn the sign of the quadratic $f(x) = x^2 - 0.75^2$ on $D=[0,1]$, where $S=[0,0.75]$ and with the unique zero contour at $\partial S=0.75$. The initial design $\mathbf{x}_{1:10}$ consists of $n=10$ inputs drawn according to Latin hypercube sampling (LHS). The observations are $Y(x) = f(x) +\epsilon$, where $\epsilon \sim t_3(0,0.1^2)$. In the top plot in Figure \ref{acquisition}, we plot the true $f(\cdot)$, the posterior mean $\hat{f}^{(100)}(\cdot)$, and associated 95\%-CI. We also show the credible band for $\partial \hat{S}$; in the respective bottom panel, we plot the acquisition functions $\im_n^{\text{MCU}}(\cdot)$, $\im_n^{\text{cSUR}}(\cdot)$, $\im_n^{\text{ICU}}(\cdot)$ and $\im_n^{\text{tMSE}}(\cdot)$ as defined in Equations (\ref{ucb}), (\ref{criterionmeesur}), (\ref{criterioneee}), and (\ref{criteriontmse}).
	
	Comparing the acquisition functions of the four criteria, we find that, besides ICU, all of the others have maxima within the shaded credible interval of the boundary $CI_{\partial S}$. In practice, we care only about the maximizer of the acquisition function, rather than its full shape, since the former drives the selection of the next sample $x_{n+1}$. The $x_{n+1}$'s selected by MCU and tMSE criteria are close. 
	For the cSUR criterion, because $\im_n^{\text{cSUR}}(x) = 0$ at $\partial \hat{S}$, there are two local maxima with a ``valley'' between them.
	The interval between the two local maxima is roughly the confidence interval $CI_{\partial S}$ for the boundary \eqref{cis}. Both MCU and tMSE select a location very close to the boundary $\hat{f}^{(n)}(x_{n+1}) \simeq 0$. We note that MCU has a flatter acquisition function, i.e., tMSE is more aggressive. In contrast, the ICU and cSUR criteria are more ``global''; in particular, ICU is the flattest among all the criteria.
	
	\begin{figure}[h]
		\centering
		\includegraphics[height=2.6in]{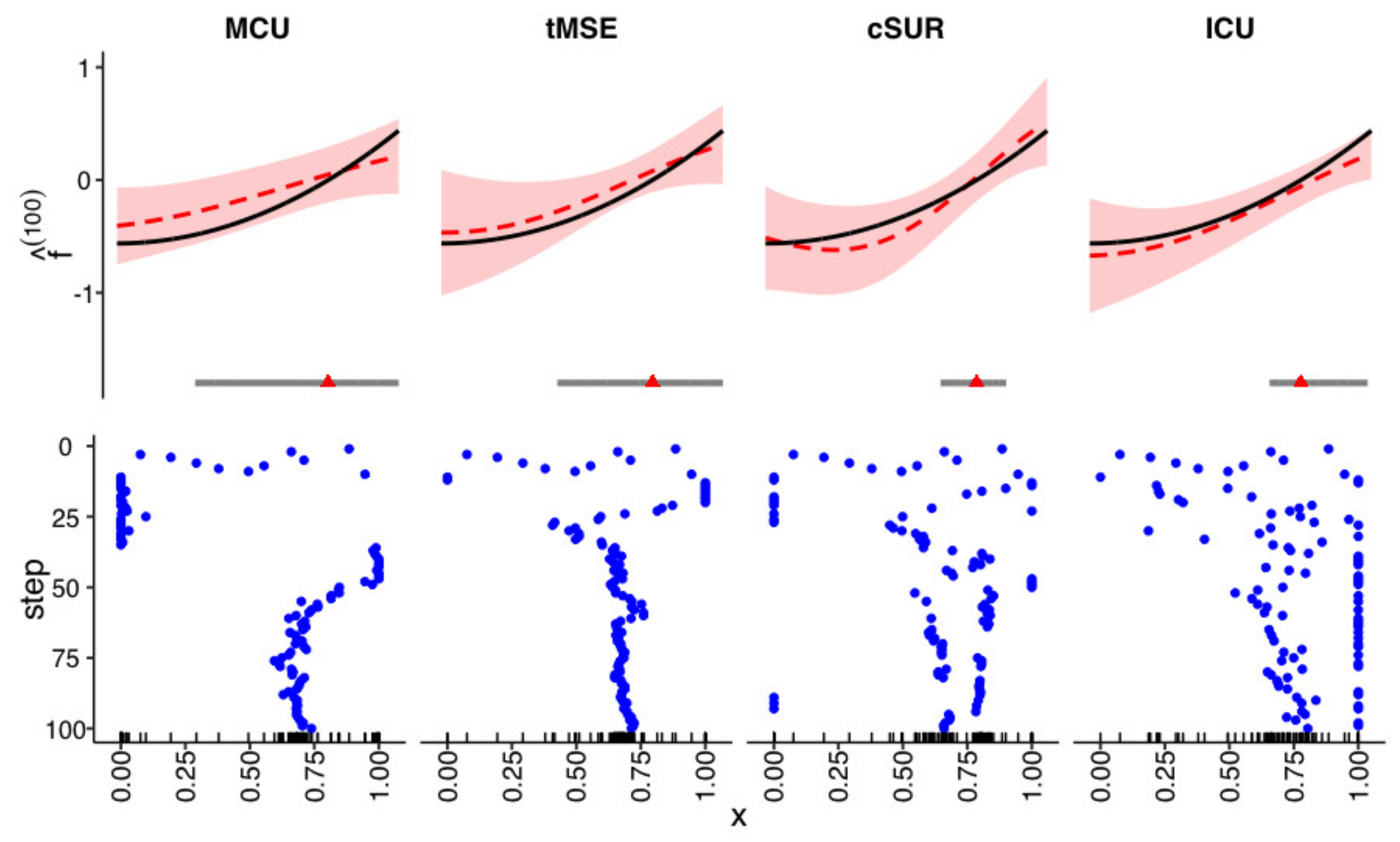}
		\caption{\emph{Top row}: Fitted metamodel $\hat{f}^{(100)}$ (dashed red line) and its 95\%-CI (shaded region) versus the true $f = (x+0.75)(x-0.75)$ (solid black), for each of the four design strategies. The estimated 95\% CI for the zero-contour $\partial S$ is marked on the $x$-axis with a grey interval; red triangle indicates the true zero-contour $\partial S=0.75$. \emph{Bottom row}: sampled inputs $x_n$ (on the $x$-axis to match the top row) as a function of step $n=1,\ldots,100$ (on the $y$-axis, moving from top to bottom) for MCU, tMSE, cSUR, and ICU criteria. The rug plots at the bottom visualize the overall distribution of $\bx$ at $n=100$. The first ten inputs are selected using a (fixed-across schemes) LHS design on $D = [0,1]$. }
		\label{designfit}
	\end{figure}
	
	After using the various acquisition functions to select $x_{n+1}$ at $n=11, \ldots, 100$, we show in Figure~\ref{designfit} the resulting designs $\bx$ and the final estimate $\hat{f}^{(100)}$ with a Gaussian observation GP metamodel. As desired, all methods target the true zero-contour at $\partial S = 0.75$. As a result, the posterior variance $s^{(n)}(x)^2$ is much lower in this neighborhood; in contrast, especially for tMSE and MCU, few samples are taken far from $x=0.75$, and the posterior uncertainty there remains high. The true zero contour is within the estimated posterior CI for all the criteria. However, the CIs for MCU and tMSE are much wider than those for the others.
	
	The bottom row in Figure~\ref{designfit} shows the  sampled location $x_{n}$ as a function of step $n$. We observe that MCU and tMSE heavily concentrate their search around the zero contour, leading to few samples  (and consequently relatively large empirical errors $\cE^{(n)}$) in other areas, although the overall error rate $\cR$ is comparable. The ICU and cSUR criteria exhibit an ``edge'' effect; that is, besides the desired zero contour $x=0.75$, multiple samples are taken close to the edges of the input space at $x=0$ and $x=1$. This occurs due to the relatively large posterior variance $s^2(\cdot)$ in those regions (which arises intrinsically with any spatial-based metamodel) that in turn strongly influences $\im^{\text{cSUR}}$ in \eqref{criterionmeesur} and $\im^{\text{ICU}}$ in \eqref{criterioneee}. Inputs sampled by the cSUR criterion bracket the contour $\partial S$ from both directions,  matching the two-hill-and-a-valley shape of $\im^{\text{cSUR}}$ in Figure~\ref{acquisition}. We note that the two sampling ``curves'' get closer as $n$ grows, indicating a gradual convergence of the estimated zero contour $\partial \hat{S}^{(n)}$, akin to a shrinking credible interval of $\hat{S}^{(n)}$.  The ICU criterion generates a much more diffuse design: it engages in more exploration and is less dependent on the current levels of the empirical error $\cE$. This eventually creates a flatter profile for $\bar{E}(x)$.
	
	The preceding discussion considered a single metamodel choice for $f$. Other metamodels will generate different design features; in particular, sensitivity to $\epsilon(x)$ will lead to a different mix of exploration ($x_n$'s far from the zero-contour) and exploitation even for the same choice of a $\im_n$ criterion. Figures~\ref{2dexpresult} and \ref{fig:2doption}, as well as Table~\ref{tbl:2d}, emphasize our message that one must jointly investigate the \emph{combinations} of $\im(\cdot)$ and $\hat{f}$ when benchmarking the ultimate performance of the algorithm.

\section{Look-Ahead Variance} \label{sec:update}
The cSUR and ICU acquisition functions $\im_n$ require estimates of the look-ahead standard deviation $s^{(n+1)}(x_*)$ conditional on sampling at $x_{n+1}=x$. A related computation is also important for efficient updating of the GP/TP metamodels during sequential design, assimilating the observation $(x_{n+1}, y_{n+1})$ into $\cA_n$. As is well known, usage of GP necessitates inverting the covariance matrix $\mathbf{K}^{-1}$ which presents a computational bottleneck as $n$ grows. Updating hinges on computing $[\mathbf{K}^{(n+1)}]^{-1}$ via applying the Woodbury identities to the current $[\mathbf{K}^{(n)}]^{-1}$.

A major advantage of the classical GP paradigm is that the posterior variance $s^{(n)}(x)^2$ is a function only of the design $\bx$; that is, it is independent of the observations $\by$. This allows an exact analytic expression for $s^{(n+1)}(x)\big|_{x_{n+1}=x}$ in terms of $x_{n+1}$.  Recall that for an existing design $\bx$, after adding a new $(x_{n+1}, y_{n+1})$, the mean and variance at location $x_*$  are updated via~\citep{chevalier2014corrected}
\begin{align}
\hat{f}^{(n+1)}_{\mathrm{Gsn}}(x_*) =& \hat{f}^{(n)}_{\mathrm{Gsn}}(x_*) + 
 \lambda^{(n)}(x_*, x_{n+1})(y^{n+1}- \hat{f}^{(n)}_{\mathrm{Gsn}}(x_{n+1})), \\
s^{(n+1)}_{\mathrm{Gsn}}(x_*)^2 =& s^{(n)}_{\mathrm{Gsn}}(x_*)^2 - 
\lambda^{(n)}(x_*, x_{n+1})^2(\taun^2+ s^{(n)}_{\mathrm{Gsn}}(x_{n+1})^2),
\end{align}
where $\lambda^{(n)}(x_*,x_{n+1})$ is a weight function 
that measures the influence of the new sample at $x_{n+1}$ on $x_*$ conditioned on the existing inputs $\bx$.

\begin{lemma}[Woodbury formula] \label{matrixinv}
	Assume $\mathbf{b}$ is a $n \times 1$ vector, $\mathbf{A}$ is a $n \times n$ matrix, and $d$ and $c$ are nonzero scalars; then we have
	\begin{align}
	[\mathbf{b}^T \quad d]
	\left[\begin{array}{cc}
	\mathbf{A} & \mathbf{b}  \\
	\mathbf{b}^T & c
	\end{array}\right]^{-1}
	\left[\begin{array}{c}
	\mathbf{b}  \\
	d
	\end{array}\right]
	= \mathbf{b}^T\mathbf{A}^{-1}\mathbf{b}-\frac{1}{c-\mathbf{b}^T\mathbf{A}^{-1}\mathbf{b}}(d-\mathbf{b}^T\mathbf{A}^{-1}\mathbf{b})^2.
	\end{align}
\end{lemma}

Using Lemma \ref{matrixinv}, we obtain the one-step-ahead variance at $x_*$:

\begin{proposition}\label{updatevarcx}
	For any $x_*$,
	\begin{align}
	\begin{split}
	\lambda^{(n)}(x_*, x_{n+1}) &= \frac{v^{(n)}_{\mathrm{Gsn}}(x_*,x_{n+1})}{\taun^2+s^{(n)}_{\mathrm{Gsn}}(x_{n+1})^2} \\
	 \Rightarrow \quad
	s^{(n+1)}_{\mathrm{Gsn}}(x_*)^2
	&= s^{(n)}_{\mathrm{Gsn}}(x_*)^2 - \frac{v^{(n)}_{\mathrm{Gsn}}(x_*,x_{n+1})^2}{\taun^2+s^{(n)}_{\mathrm{Gsn}}(x_{n+1})^2}.
	\end{split}
	\label{updategv}
	\end{align}
	In particular, after sampling at $x_{n+1}$ the local updated posterior variance is proportional to the current $s^{(n)}_{\mathrm{Gsn}}(x_{n+1})^2$ with a proportionality factor \citep{hu2015sequential}:
	\begin{align}
	\frac{s^{(n+1)}_{\mathrm{Gsn}}(x_{n+1})^2}{s^{(n)}_{\mathrm{Gsn}}(x_{n+1})^2} &= \frac{\taun^2}{\taun^2+s^{(n)}_{\mathrm{Gsn}}(x_{n+1})^2}. \label{updatevg}
	\end{align}

\end{proposition}
The above lemma is our basis for calculating the acquisition function for the cSUR criterion (\ref{criterionmeesur}) that requires only \eqref{updatevg} and the ICU criterion (\ref{criterioneee}). As we see below, because \eqref{updategv} holds only in the Gaussian prior/Gaussian likelihood setting, further approximations are required to apply \eqref{updatevarcx}--\eqref{updatevg} for the alternative metamodels. Such look-ahead variance expressions are of independent interest, applicable beyond the context of level set estimation.

 A limitation of using a non-Gaussian observation  or classification likelihood is that, unlike for Gaussian observation GP, there are no exact variance look-ahead formulas for the resulting $t$-GP, Cl-GP and TP metamodels. There are two main reasons for this. First, both the posterior mean $\hat{f}^{(n+1)}(x_*)$ in (\ref{meant}) and \eqref{meanz} and the posterior variance $s^{(n+1)}(x_*)^2$ in (\ref{covt})  and \eqref{covz} for $t$-GP and Cl-GP depend on the posterior mode  $\tilde{\mathbf{f}}_{t\mathrm{GP}}^{(n+1)}$ or $\tilde{\mathbf{z}}_{Cl}^{(n+1)}$, which changes every step. Therefore, they cannot be accessed in advance. Furthermore,  for $t$-GP and Cl-GP $s^{(n+1)}(x_*)$ depends on the next-step Hessian $\bm{W}$ (namely on $w_{n+1}^{(n+1)}$), and for TP  $s^{(n+1)}(x_*)$ depends on $\beta^{(n+1)}$. Both of
	these again depend on $y_{n+1}$. To overcome this challenge, we develop an approximation $\hat{s}^{(n+1)}(\cdot)$ for each metamodel. Our strategy is to replace each inaccessible term with its expected value from the point of view of step $n$. For example, we calculate the expectation of $\tilde{\mathbf{f}}_{t\mathrm{GP}}^{(n+1)}$, $\tilde{\mathbf{z}}_{Cl}^{(n+1)}$ and $\beta^{(n+1)}$ with respect to $\mathcal{A}_n$. Propositions~\ref{updatevarcxt}-\ref{updatevarcxcl}-\ref{updatevarcxtp} provide the resulting look-ahead formulas for $t$-GP, Cl-GP and TP  respectively, with derivation details in Appendix~\ref{app:lookahead}.
	
	\begin{proposition}\label{updatevarcxt}
		For any $x_*$, the formula for the look-ahead variance for $t$-GP is
		\begin{align}
		\hat{s}^{(n+1)}_{t\mathrm{GP}}(x_*)^2
		& := s^{(n)}_{t\mathrm{GP}}(x_*)^2 - \frac{v^{(n)}_{t\mathrm{GP}}(x_*,x_{n+1})^2}{(\tau^2 \frac{\nu+1}{\nu-1})+s^{(n)}_{t\mathrm{GP}}(x_{n+1})^2}. \label{updatetv}
		\end{align}
	\end{proposition}
	
	\begin{proposition}\label{updatevarcxcl}
		Let $\check{v}_{n+1} = v_{n+1}^+p_++v_{n+1}^-p_-$, where
		\begin{align}
		v_{n+1}^+ &= \frac{\phi(\hat{\z}_{\mathrm{Cl}}^{(n)}(x_{n+1}))^2}{\Phi(\hat{\z}_{\mathrm{Cl}}^{(n)}(x_{n+1}))^2} + \frac{\hat{\z}_{\mathrm{Cl}}^{(n)}(x_{n+1})\phi(\hat{\z}_{\mathrm{Cl}}^{(n)}(x_{n+1}))}{\Phi(\hat{\z}_{\mathrm{Cl}}^{(n)}(x_{n+1}))}, \\
		p_+ &=  \Phi\bigg(\frac{\hat{\z}^{(n)}(x_{n+1})}{\sqrt{1+s^{(n)}_C(x_{n+1})^2}}\bigg),
		\end{align}
		\text{and}
		\begin{align}
		v_{n+1}^- &= \frac{\phi(\hat{\z}_{\mathrm{Cl}}^{(n)}(x_{n+1}))^2}{\Phi(-\hat{\z}_{\mathrm{Cl}}^{(n)}(x_{n+1}))^2} -\frac{\hat{\z}_{\mathrm{Cl}}^{(n)}(x_{n+1})\phi(\hat{\z}_{\mathrm{Cl}}^{(n)}(x_{n+1}))}{\Phi(-\hat{\z}_{\mathrm{Cl}}^{(n)}(x_{n+1}))},  \\
		p_- &= 1 - p_+.
		\end{align}
		For any $x_*$, the formula for the look-ahead variance for Cl-GP is
		\begin{align}
		\hat{s}^{(n+1)}_{\mathrm{Cl}}(x_*)^2 :=  {s^{(n)}_{\mathrm{Cl}}(x_*)^2} - \frac{v^{(n)}_{\mathrm{Cl}}(x_*,x_{n+1})^2}{ (\check{v}_{n+1})^{-1}+s^{(n)}_{\mathrm{Cl}}(x_{n+1})^2}.
		\end{align}
	\end{proposition}
	
	\begin{proposition}\label{updatevarcxtp}
		For any $x_*$, the formula for the look-ahead variance for TP is
		\begin{align}
		s^{(n+1)}_{\mathrm{TP}}(x_*)^2 &= \frac{\nu + \check{\beta}^{(n+1)} - 2}{\nu + n - 1} s^{(n+1)}_{\mathrm{Gsn}}(x_*)^2, \label{updatetpv}
		\end{align}
		where $\check{\beta}^{(n+1)} =\beta^{(n)} + \frac{\nu}{\nu - 2}.$
	\end{proposition}
	
We note that in our experiments we only use the above to evaluate $\im_n$, and  directly re-estimate $\tilde{f}^{(n+1)}$ at each step of the sequential design.

\section{Synthetic Experiments} \label{sec:synthetic}

\subsection{Benchmark Construction} \label{benchmark}

As synthetic experiments, we consider three benchmark problems in dimension $d=1, 2$, and 6. For the latter two we employ the widely used \emph{Branin-Hoo} 2-D and \emph{Hartman} 6-D functions; see, for example, \cite{picheny2013benchmark}. The original functions have been rescaled to map their sample space $D$ onto $[0,1]^d$; see Table~\ref{syntheticfunc}.

The latent functions are chosen to cover a variety of problem properties. The quadratic $f$ in 1-D is strictly monotonically increasing, yielding a single boundary $\partial S$. The original Branin-Hoo function \citep{picheny2013benchmark} is modified so that $f$ is increasing in $x^1$ and the zero-level set has a non-trivial shape in $x^2$. The \emph{Hartman} is a multimodal function with a complex zero contour. The parameters in the original \emph{Hartman} function described in \cite{picheny2013benchmark} are adjusted to reduce the "bumps" in the zero contour and make the problem more appropriate for the sign classification task.

\begin{table*}[ht]
	\caption{Response surfaces ${x} \mapsto f({x})$ for synthetic experiments.}
	\centering
	\begin{tabular}{ll}
		\hline\noalign{\smallskip}
		Quadratic (1-D) & $f(x) = (x+0.75)(x-0.75)$
		 with $x\in[0,1]$\\
		\noalign{\smallskip}\hline\noalign{\smallskip}
		Branin-Hoo (2-D) & $f({x}) = \frac{1}{178} \big[\big(\bar{x}^1-\frac{5.1(\bar x^2)^2}{4\pi^2}+\frac{5\bar x^2}{\pi}-20\big)^2 + (10-\frac{10}{8\pi})\cos(\bar x^1)-181.47\big] $\\
		& with: $\bar x^1 = 15x^1$, $\bar x^2 = 15x^2 - 5$, $x^1, x^2 \in [0,1]$\\
		\noalign{\smallskip}\hline\noalign{\smallskip}
		Hartman6 (6-D) & $f({x}) = \frac{-1}{0.1}\big[\sum_{i=1}^{4}C_i\exp\big(-\sum_{j=1}^{6}a_{ji}(x^j-p_{ji})^2\big)-0.1\big]$\\
		& with: $\mathbf{C} = [0.2,0.22,0.28,0.3]$ \\
		& $\mathbf{a} = \begin{bmatrix}
		
		8.00 & 0.50 & 3.00 & 10.00 \\
		3.00 & 8.00 & 3.50 & 6.00 \\
		10.00 & 10.00 & 1.70 & 0.50\\
		3.50 & 1.00 & 8.00 & 8.00 \\
		1.70 & 6.00 & 10.00 & 1.00 \\
		6.00 & 9.00 & 6.00 & 9.00 \end{bmatrix}\!, \quad
	\mathbf{p} = \frac{1}{10^{4}}\begin{bmatrix}
		
		1312 & 2329 & 2348 & 4047 \\
		1696 & 4135 & 1451 & 8828 \\
		5569 & 8307 & 3522 & 8732 \\
		124 & 3736 & 2883 & 5743 \\
		8283 & 1004 & 3047 & 1091 \\
		5886 & 9991 & 6650 & 381 \end{bmatrix}$ \\
	\noalign{\smallskip}\hline
	\end{tabular}
	\label{syntheticfunc}
	\end{table*}
	
	A large number of factors can influence the performance of metamodels and designs. In line with the stochastic simulation perspective, we concentrate on the impact of the simulation noise and consider four observation setups. These cover a variety of noise distributions and signal-to-noise ratio,  measured through the proportion of standard deviation $\sigma_\tau$ to the range $R_f$ of the response. The first two settings use
	\emph{Student-$t$} distributed noise, with (i) low $\sigma_\tau$ and (ii) high $\sigma_\tau$. The third setting uses (iii) Gaussian mixture noise to further test misspecification of $\epsilon$. The fourth setting considers the challenging case of (iv) a heteroscedastic Student-$t$ noise with state-dependent degrees of freedom. In total we have $3 \times 4 \times 4 \times 6$ experiments (indexed by their dimensionality, noise setting, design heuristic, and metamodel type).
	
	Besides the noise distribution, we fix all other metamodeling aspects. All schemes are initialized with $n_0=10d$ inputs drawn from an LHS design on $[0,1]^d$ and use the SE kernel \eqref{covf} for the covariance matrix $\bm{K}$. To analyze for the variability due to the initial design and the noise realizations, we perform 100 macroruns of each design/acquisition function combination. For each run, the same initial inputs are used across all GP metamodels and designs, but otherwise the initial $\mathbf{x}_{1:n_0}$ vary across runs.

	\begin{table*}[ht]
		\caption{Stochastic simulation setup for synthetic experiments.  ($R_f \equiv \max_{{x}} f({x})-\min_{{x}} f({x}) = 1$) 
		}
		\centering
		\begin{tabular}{ll}
			\hline\noalign{\smallskip}
			Initial design  & Latin hypercube sampling of size $n_0 = 10d$\\
			Total budget $n$ & $d = 1, n = 100; \quad d = 2, n = 150; \quad d = 6, n = 1000$ \\
			Test set size $M=|\cD|$ & $d = 1, M = 1000; \quad d = 2, M = 500; \quad d = 6, M = 1000$  \\ \noalign{\smallskip}\hline\noalign{\smallskip}
			Noise setting for $\epsilon(x)$ & (i) $t/ \text{small}:t_3(0, (0.1R_f)^2)$ \\
			& (ii) $t/ \text{large}: t_3(0, (0.5R_f)^2)$ \\
			& (iii) $Gsn/ \text{mix: 50/50 mix of }\; \mathcal{N}(0,(0.5R_f)^2) \text{ and }\; \mathcal{N}(0,R_f^2)$ \\
			& (iv) $t/ \text{hetero}: t_{6-4x^1}(0,(0.4(4x^1+1))^2)$  \\
			\noalign{\smallskip}\hline
		\end{tabular}
		\label{syntheticnoise}
	\end{table*}

	\textbf{Optimization of the Improvement Metric:} We employed the MCU, ICU, tMSE and cSUR criteria to maximize the improvement metric $\mathcal{I}$ and select the next input $x_{n+1}$. This maximization task is nontrivial in higher dimensions because $\im$ is frequently multimodal and can be flat around its local maxima. We use a genetic optimization approach as implemented in the \texttt{ga} function in MATLAB, with tolerance of $10^{-3}$ and $200$ generations. This is a global, gradient-free optimizer that uses an evolutionary algorithm to explore the input space $D$. 
	
	\textbf{Evaluation of Performance Metrics:}
	Recall that evaluating the quality of $\partial \hat{S}$ is based on $\cR$ and $\cE$ from \eqref{erc} and \eqref{eec} that require integration over $D$. In practice, these are computed based on a weighted sum over a finite $\mathcal{D}$, $\hat{\cE} := \sum_{m=1}^{M} \Phi\big(\frac{-|\hat{f}(x_m)|}{s(x_m)}\big) \mu(x_m)$ for a space-filling sequence $\mathcal{D} \equiv x_{1:M} \in D$ of test points. In 1-D experiments $\cD$ was an equispaced grid  of size $M=1000$. In higher dimensions, to avoid the use of a lot of test points that are required to ensure an accurate approximation, we adaptively pick $\mathcal{D}$ that targets the critical region close to the zero contour.  To do so, we replace the integral with a weighted sum:
	\begin{align}
	\begin{split}
	\cR \simeq&  \frac{p_c}{M_1} \sum_{x_{1:M_1} \in D_1} \!\mathbb{I} (\sgn f(x_m)\neq \sgn \hat{f}(x_m)) 
	 + \frac{(1-p_c)}{M_2} \sum_{x_{1:M_2} \in D_2} \mathbb{I} (\sgn f(x_m)\neq \sgn \hat{f}(x_m)),
	\end{split}
	\end{align}
	where $M = M_1 + M_2$ and the test locations $x_{1:M_1}$ and $x_{1:M_2}$ are subsampled from a large space-filling (scrambled Sobol) sequence on $D$. The weight $p_c$ determines the relative volume of $D_1$ and $D_2 = D \backslash D_1$, where on $D_1 = \{ x : f(x) \simeq 0 \}$ we are close to the zero contour.
	In the experiments below we use $M_1 = 0.8M, M_2 = 0.2M$, and $p_c = 0.4$, so that the density of test points close to $\partial S$ is double relative to those far from the zero contour. We employ the same strategy for speeding the evaluation of  the empirical error $\cE$.
	
	\textbf{Surrogate Inference:} Values of hyperparameters $\bm{\vartheta}$ are crucial for good performance of GP metamodels. We estimate $\bm{\vartheta}$ using maximum likelihood. Except for TP, all models are fitted  with the open source package \texttt{GPstuff} \citep{vanhatalo2012bayesian} in MATLAB. TPs are fitted with the \texttt{hetGP}~\citep{binois2016practical} package in \texttt{R}. Auxiliary tests did not reveal any significant effects from using other available tools for plain GPs and $t$-GP, such as \texttt{GPML} \citep{rasmussen2010gaussian}.

	In principle, the hyperparameters $\bm{\vartheta}$ change at every step of the sequential design, in other words, whenever $\mathcal{A}_n$ is augmented with $(x_{n+1}, y_{n+1})$. To save time however, we do not update $\bm{\vartheta}$ at each step. Instead, we first estimate the hyperparameters $\bm{\vartheta}$ based on the initial design $\mathcal{A}_{n_0}$ and then freeze them, updating their values only every few steps. Specifically, $\bm{\vartheta}$ is re-estimated at steps $n_0+1, n_0+2, n_0+4, n_0+8, n_0+16, \ldots $ (as the sample size becomes large, the inference of hyperparameters becomes more stable).
	
	The lengthscales $\theta_i$ are the most significant for surrogate goodness of fit. A too-small lengthscale will make the estimated  $\hat{f}$ look ``wiggly'' and might lead to overfitting, while $\theta_i$ too large will fail to capture an informative shape of the true $f$ and hence $S$. Since our input domain is always $[0,1]^d$, we restrict  $\theta_i \in [0.3, 2]~\forall i$ to be on the order of the length of the sample space $D$.
	
	\textbf{Computational Overhead:} All the considered metamodels are computationally more demanding than the baseline Gaussian GP. For $t$-GP and Cl-GP, additional cost arises due to the Laplace approximation.  TP necessitates estimation of the parameter $\nu$ and also the computation of $\beta$ in \eqref{TPcov}. In the experiments considered, the respective computation times were roughly double to triple relative to the Gaussian GP. In terms of sequential design, MCU, tMSE, and cSUR have approximately equal overhead; ICU is significantly more expensive because it requires evaluating the sum in \eqref{criterioneee}. Note that all heuristics include two expensive steps: optimization for $x_{n+1}$ and computation of $\hat{f}^{(n)}$ and $s^{(n)}$ (and/or $\hat{s}^{(n+1)}$). 
	
	Overall timing of the schemes is complicated because of the combined effects of $n$ (design budget), $M$ (size of test set), and the use of different software (some schemes run in \texttt{R} and others in Matlab). Most important, the ultimate computation time is driven by the simulation cost of generating $Y(x)$-samples, which is trivial in the synthetic experiments but assumed to be large in the motivating context.

	\subsection{Comparison of GP Metamodels} \label{compgp}

\begin{figure*}[ht]
		\begin{center}
			\includegraphics[width=0.95\textwidth,trim=0.3in 0.4in 0.3in 0.2in]{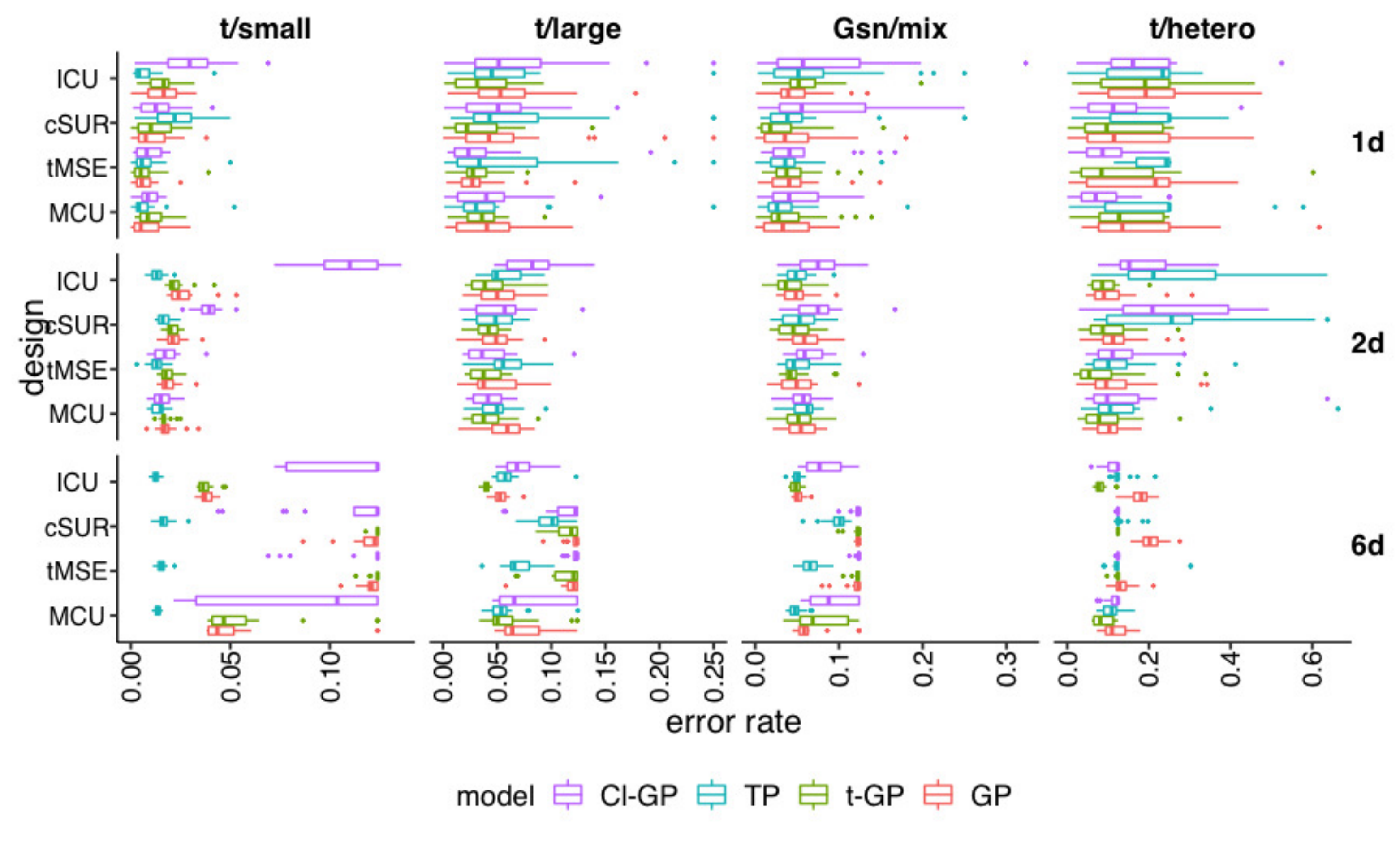}
		\end{center}
		\caption{Boxplots of final error rate $\mathcal{ER}^{(n)}$ from \eqref{erc} across designs (rows) and noise setups (columns). Colors correspond to different GP metamodels. Note that $x$-axis limits are different across columns. Top row is for the 1-D experiment and design size $n=100$; middle row:  2-D Branin-Hoo function with $n=150$; bottom row: 6-D Hartman6 function with $n=1000$.}
		\label{fig:r}
	\end{figure*}

	Figure \ref{fig:r} shows the boxplots of the error rate $\cal{ER}$ of $\hat{S}^{(N)}$ at the final design ($N=100$ in 1-D; $N=150$ in 2-D; $N=1000$ in 6-D). The plots are sorted by noise settings and design strategies, facilitating comparison between the discussed metamodels.  In Table \ref{tbl:2d}, we list the best metamodel and design combination in each case. Several high-level observations can be made. First, we observe the limitations of the baseline Gaussian GP metamodel, which cannot tolerate too much model misspecification. As the noise structure gets more complex, the classical GP surrogate begins to show increasing strain; in the last $t/hetero$ setup, it is both unstable (widely varying performance across runs) and inaccurate, with error rates upward of 30\% on ``bad'' runs. In addition, according to results shown in Table \ref{tbl:2d}, across all of the twelve cases, besides 1d example with $t/{small}$ noise, the Gaussian GP never performs as the best model. This result is not surprising but confirms that the noise distribution is key for the contour-finding task and illustrates the nonrobustness of the Gaussian observation model, due to which outliers strongly influence the inference.
	
	Second, we document that the simple adjustment of using Student-$t$ observations significantly mitigates the above issue. $t$-GP performs consistently and significantly better than Gaussian GP in essentially all settings. This result is true even when both models are misspecified (the $Gsn/mix$ and $t/hetero$ cases). The performance of $t$-GP was still better (though not statistically significantly so) when we tested it in the setting of homoscedastic Gaussian noise (not shown in the plots). The latter fact is not surprising---$t$-GP adaptively learns the degrees-of-freedom parameter $\nu$ and hence can ``detect'' Gaussian noise by setting $\nu$ to be large. Conversely, in heavy-tailed noise cases, the use of $t$ samples will effectively ignore outliers \citep{o1979outlier} and thus produce more accurate predictions than working with a Gaussian observation assumption. We find that $t$-GP can handle complex noise structures and offers a good choice for all-around performance, making it a good default selection for applications. It brings smaller error rate $\cal{ER}$, more stable hyperparameter estimation, less contour bias, and tighter contour CI. Moreover $t$-GP is significantly better than all the other GPs in eight of the twelve setups, indicating that $t$-GP is essentially the best out of all GP metamodels in most cases.
	
	Third, we also inspect the performance of the TP metamodel. As shown in Table \ref{tbl:2d}, TP is the best in two cases out of the twelve, both of which are with the $t /{small}$ noise. We note that TP works worst in $t/{hetero}$ cases, having both large error rate $\cal{ER}$ and empirical error $\cE$. Therefore, TP does not work well in cases with low signal-to-noise ratio or greatly misspecified noise. This may be related to the parameterization of TPs, with noise handled in the kernel, which seems less robust to misspecification. Also, since TPs revert to GPs as $n$ increases, the advantage of flexibility offered by the modeling decreases as iterations goes and thus may not last enough for low signal-to-noise ratios, which require more samples. It is apparent for instance in Figure \ref{fig:eeerstep}, where the error at step 0 is lower than for its counterparts.

	Fourth, Cl-GP is also better than Gaussian GP in some cases with tMSE and MCU designs (except for the 6-D cases, where the error rate $\cal{ER}$ of MCU is not significantly different from that of ICU, although mean of ICU is slightly smaller). There is significant improvement for models with low signal-to-noise ratio; the only exception is for the low-noise setup where Cl-GP underperforms classical GP. This matches the intuition that employing classification ``flattens'' the signal by removing outliers. By considering only the sign of the response, the classification model disregards its magnitude, simplifying the noise structure at the cost of some information loss. The net effect is helpful when the noise is mis-specified or too strong so as to interfere with learning the mean response. It is detrimental if the above gain is outweighed by the information loss, as apparently happens in the 6-D experiments. Of note, Cl-GP with MCU design has the smallest error rate among all models in one ($t/hetero$ in 1-D) out of 12 cases shown in Table~\ref{tbl:2d}. We also observe, however, that the stability of Cl-GP is highly dependent on the design: some designs create large across-run variations in performance. We hypothesize that this is due to a more complex procedure for learning the hyperparameters of Cl-GP; therefore, designs that are not aggressive enough to explore the zero contour region (such as cSUR) face difficulties in estimating $\vartheta$. As a result, relative to $t$-GP, Cl-GP has  larger sampling variances.

	\begin{table*}[htb]
		\caption{Mean (w/standard deviation) error rate $\cR$ and corresponding best-performing sequential design heuristic for the 1-D, 2-D, and 6-D synthetic case studies. Results are based on 100 macroreplications of each scheme.}
		\centering
		{\small 		\begin{tabular}{lrlrlrlrl}
				\hline\noalign{\smallskip}
				Model  & \multicolumn{2}{c}{ $t/{small}$}  & \multicolumn{2}{c}{ $t/large$} & \multicolumn{2}{c}{$Gsn/{mix}$} & \multicolumn{2}{c}{$t/{hetero}$} \\
				\noalign{\smallskip}\hline\noalign{\smallskip}
				& \multicolumn{8}{c}{ 1-D \emph{Quadratic} } \\
				\noalign{\smallskip}\hline\noalign{\smallskip}
				GP & \emph{tMSE} & 0.73\% (0.60\%)  & tMSE & 3.24\% (2.79\%) & MCU & 3.87\% (3.17\%) & cSUR & 15.68\% (12.15\%) \\
				$t$-GP & tMSE & 0.80\% (0.93\%) & \emph{tMSE} & 3.15\% (1.83\%) & \emph{cSUR} & 3.28\% (3.74\%) & cSUR & 12.50\% (9.05\%) \\
				TP  & MCU & 0.97\% (0.84\%) & MCU & 5.93\% (5.60\%) & tMSE & 5.09\% (4.40\%) & ICU & 16.44\% (10.14\%)\\
				Cl-GP & tMSE & 0.87\% (0.64\%) & tMSE & 3.39\% (4.16\%) & MCU & 4.99\% (3.77\%) & \emph{MCU} & 8.83\% (7.35\%) \\
				\noalign{\smallskip}\hline\noalign{\smallskip}
				& \multicolumn{8}{c}{ 2-D \emph{Branin-Hoo} } \\
				\noalign{\smallskip}\hline\noalign{\smallskip}
				GP & MCU &1.78\% (0.57\%) & cSUR & 4.75\% (1.95\%) & ICU & 4.92\% (1.86\%) & MCU & 10.36 \% (3.94\%) \\
				$t$-GP & MCU &1.70\% (0.29\%) & \emph{tMSE} &3.95\% (1.47\%) & \emph{ICU} &4.10\% (2.07\%) & \emph{tMSE} & 9.00\% (8.66\%) \\
				TP & \emph{tMSE} & 1.27\% (0.41\%) & MCU & 4.79\% (1.84\%) & ICU & 5.19\% (1.68\%) & MCU & 12.75 \% (9.02\%)\\
				Cl-GP & MCU & 1.56\% (0.51\%) & MCU &4.27\% (1.59\%) & MCU & 5.71\% (1.85\%) & tMSE &13.23\% (7.74\%) \\
				\noalign{\smallskip}\hline\noalign{\smallskip}
				& \multicolumn{8}{c}{ 6-D \emph{Hartman6} } \\
				\noalign{\smallskip}\hline\noalign{\smallskip}
				GP & ICU &3.81\% (0.34\%)  & ICU &5.33\% (0.54\%) & ICU & 5.19\% (0.70\%) & MCU &11.67\% (2.89\%)\\
				$t$-GP & ICU &3.75\% (0.40\%) & \emph{ICU} &3.98\% (0.47\%)  & \emph{ICU} & 4.86\% (0.67\%) & \emph{ICU} &8.25\% (1.60\%) \\
				TP  & \emph{ICU} &1.25\% (0.20\%) & MCU &5.66\% (1.98\%) & MCU &4.88\% (0.88\%) & MCU &10.69\% (2.34\%)\\
				Cl-GP & MCU &7.99\% (4.69\%) & ICU &7.20\% (0.66\%) & ICU & 8.31\% (2.44\%) & ICU &11.11\% (2.20\%) \\
				\noalign{\smallskip}\hline
		\end{tabular}}
		\label{tbl:2d}
	\end{table*}

	\subsection{Empirical Errors and Uncertainty Quantification}

\begin{figure*}[htb]
		\begin{center}
			\includegraphics[width=0.95\textwidth,trim=0.3in 0.4in 0.3in 0.2in]{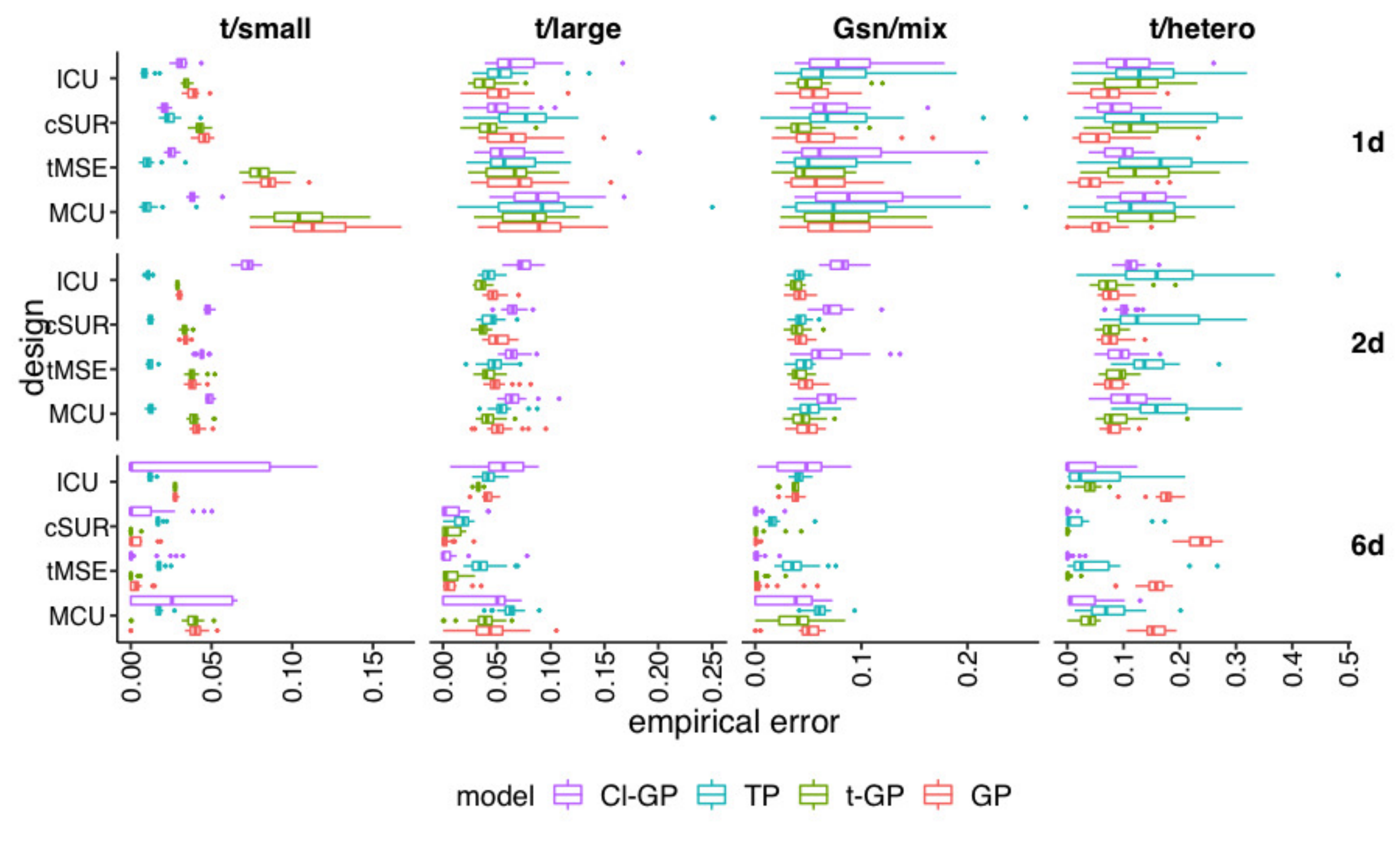}
		\end{center}
		\caption{Empirical error $\cE^{(n)}$ in Eq.~(\ref{bci}) for GP, $t$-GP, TP, Cl- GP, and MCl-GP metamodels (colors), using MCU, tMSE, cSUR and ICU-based designs (sub rows) with $n=100$ in 1-D, $n=150$ in 2-D, and $n=1000$ in the 6-D experiments (rows).}
		\label{fig:e}
	\end{figure*}

	Figure~\ref{fig:e} shows the empirical errors $\cE$ that are supposed to proxy the true error rates $\cal{ER}$. Overall, we find that MCU tends to produce the largest $\cE$, and ICU the smallest. These results are consistent with their design construction and local behavior: MCU heavily concentrates around $\partial \hat{S}$, which leads to little information collected about other regions, especially around the boundaries of sample space $D$ and hence relatively large $\bar{E}(x)$ there, inflating $\cE$. Conversely, the objective function of ICU is precisely the myopic minimization of $\cE_{n+1}$. The other two designs are intermediate versions in terms of minimizing $\cE$. The tMSE heuristic tends to target the zero contour plus the edges of $D$, while cSUR tends to broadly target a ``credible band'' around $\partial \hat{S}$. Both approaches are better at reducing $\cE$ compared with MCU but are not directly aimed at this. This logic is less consistent for the classification models, where tMSE often yields  the lowest $\cE$. This result echoes  Section \ref{compgp}, namely, that classification GPs tend to perform better with MCU and tMSE designs in lower dimensional cases. TPs tend to have a greater 
	empirical error $\cE$ when the noise is misspecified or in higher dimensional experiments, consistent with the conclusions obtained regarding the error rate $\mathcal{ER}$. 
	
\begin{figure*}[ht]
		\begin{center}
			\includegraphics[width=0.95\textwidth,trim=0.2in 0.3in 0.1in 0.2in]{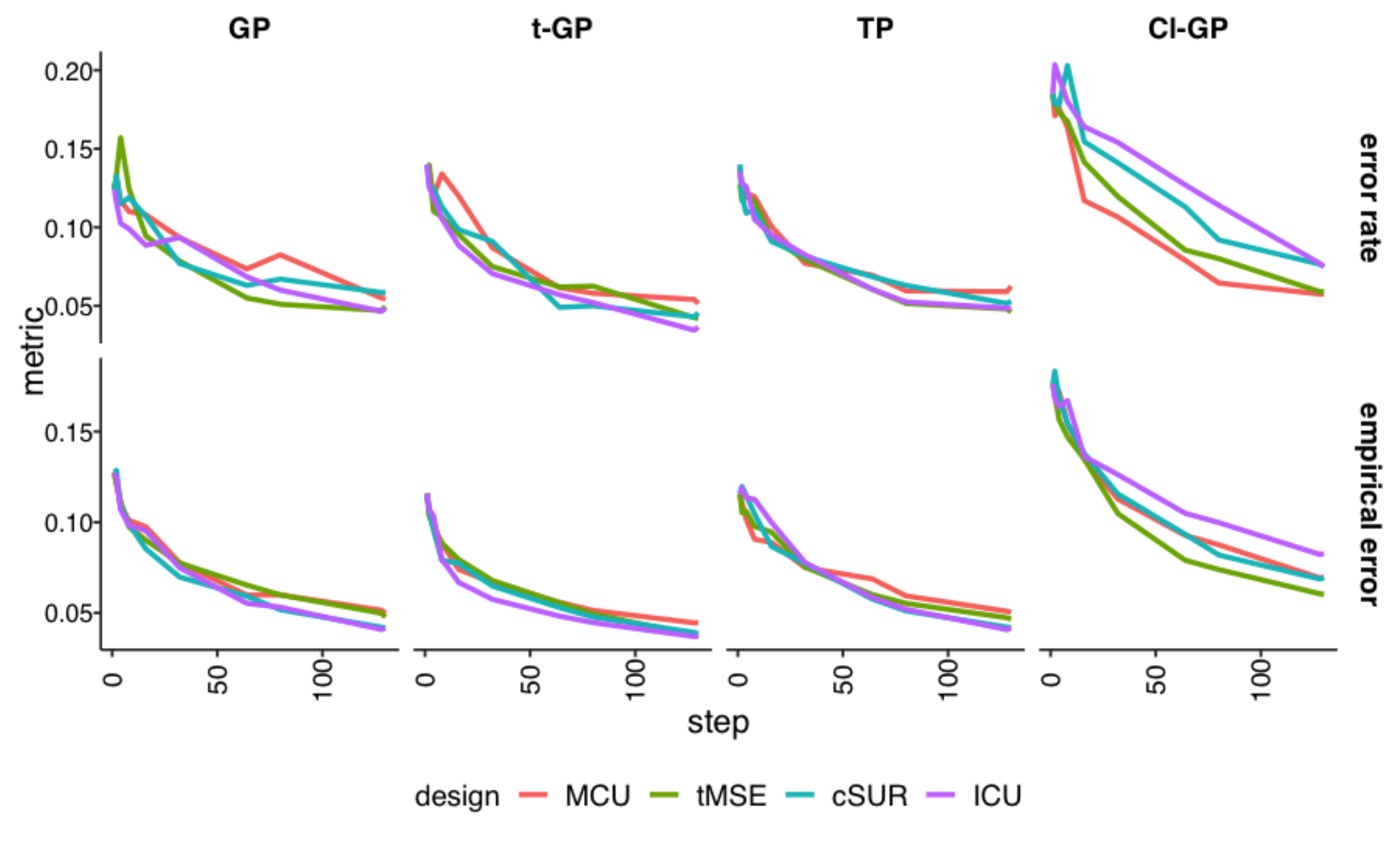}
		\end{center}
		\caption{Error rate $\cR^{(n)}$ \eqref{erc} and surrogate-based uncertainty measure $\cE^{(n)}$ \eqref{eec} as a function of step $n$ in the 2-D $Gsn/{mix}$ setting. We compare six metamodels (columns) and four DoE's (colors). The $y$-axis limits differ across rows. We plot median results across 20 macroreplications of each scheme. }
		\label{fig:eeerstep}
	\end{figure*}

	As a further visualization, Figure \ref{fig:eeerstep} shows the median error rate $\cal{ER}$ (\ref{erc}) and empirical error $\cE$ in Eq.~(\ref{eec}) as a function of step $n$ in the 2-D $Gsn/mix$ experiments. This illustrates the learning rates of different schemes as data is collected and offers a further comparison between the true $\cal{ER}$ and the self-reported $\cE$ of the same scheme.
	We observe that some metamodels underperform for very low $n$, even if they eventually ``catch up'' after sufficiently large simulation budget. This is especially pronounced for the classification Cl-GP metamodel, which yields very high $\mathcal{ER}^{(n)}$ (which is also much higher than the self-reported $\cE$) for $n$ small. We also note that Cl-GP appears to enjoy faster reduction in $\mathcal{ER}^{(n)}$ compared with the baseline Gaussian GP, which we conjecture is due to better resistance against $Y$-outliers that distract plain GP's inference of $S$. Comparing the two rows of the figure, we note that discrepancies between $\mathcal{ER}$ and $\cE$ correlate with degraded performance, namely, the metamodel being unable to properly learn the response surface is reflected in  poor uncertainty quantification. Moreover, the results suggest that the wedge in performance of different design criteria tends to persist; for example MCU and ICU frequently have not only the highest/lowest $\cE^{(n)}$ but also the slowest/fastest rate of \emph{reduction} in $\cE^{(n)}$ as $n$ grows. Consistent with results in Section \ref{compgp}, Cl-GP with ICU criterion yields both greater error rate $\cal{ER}$ and empirical error $\cE$ in 2-D experiments.

	\subsection{Designs for Contour Finding} \label{optdesign}
	
\begin{figure*}[htb]
		\begin{center}
			\includegraphics[width=0.95\textwidth,trim=0.3in 0.2in 0.3in 0.2in]{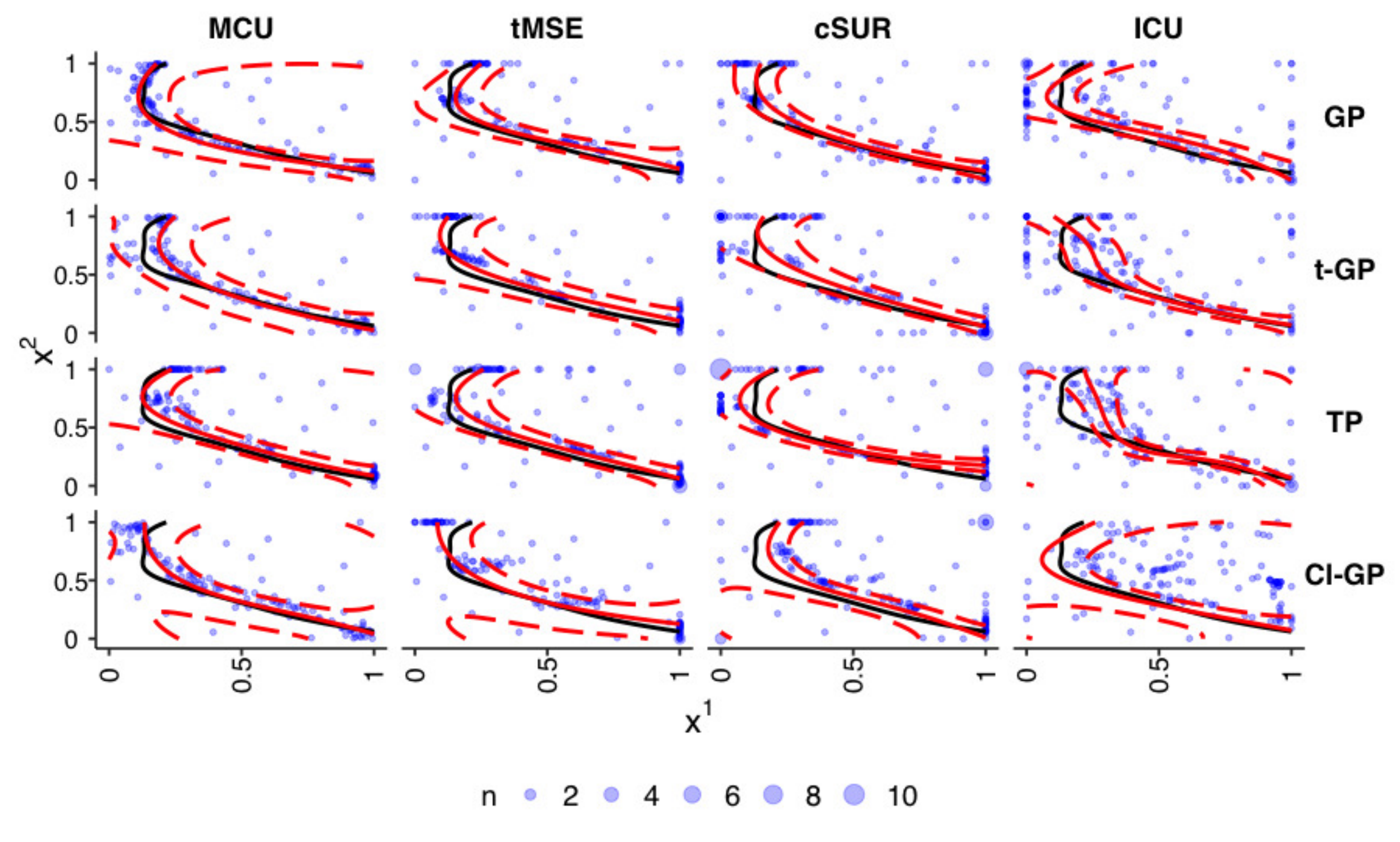}
		\end{center}
		\caption{Estimates of the zero contour $\partial \hat{S}$ for the 2-D Branin-Hoo example with $t/{large}$ noise setting. We show $\partial \hat{S}^{(n)}$ (red solid line) at step $n=150$, with its 95\% credible band (red dashed lines), the true zero contour $\partial S$ (black solid line) and the sampled inputs $\mathbf{x}_{1:n}$ (replicates indicated with larger symbols). We compare across the six metamodels (rows) and four DoE heuristics (columns).}
		\label{2dexpresult}
	\end{figure*}

	A key goal of our study is qualitative insights about experimental designs most appropriate for noisy level set estimation. Through identifying the best-performing heuristics we get an inkling regarding the structure of near-optimal designs for \eqref{eq:objective}. In this section we illustrate the latter within a 2-D setup that can be conveniently visualized.  Taking the $t/{large}$ experiment as an example, in Figure \ref{2dexpresult} we plot the fitted zero contour $\partial \hat{S}$ at $N=150$ together with the chosen inputs $\mathbf{x}_{1:150}$ across the six metamodels and the four $\im$ heuristics. As expected, most of the designs are around the contour $\partial S$, which is the intuitive approach to minimize the error $\cal{ER}$. Nevertheless, we observe significant differences in designs produced by different $\im$'s. The MCU criterion places most of the samples close to the estimated zero contour $\partial \hat{S}$, reflecting its aggressive exploitation nature. For tMSE, the samples tend to cluster at several subregions of $\partial \hat{S}$ and on the edges of $D$. For cSUR, $\mathbf{x}_{1:n}$ cover a band along $\partial \hat{S}$, resembling the shape of the MCU design but more dispersed. For ICU the design is much more exploratory, covering a large swath of $D$. All these findings echo the 1-D setting in Figure~\ref{designfit}.

	One feature we observe is a so-called edge effect, that is, designs that focus on the edges of the input space. This effect arises due to the intrinsically high posterior uncertainty $s(x)$ for $x$ around $\partial D$. It features strongly in tMSE and cSUR (which have about 45\% of the inputs along the edge) and to some extent in ICU (about 30\% of inputs in this example). In contrast, MCU strongly discounts any region that is far from $\partial \hat{S}$. In the given 2-D experiment, we obtain some inputs directly on the boundary $\partial D = \{ x^1 \in \{0,1\} \cup \{x^2 \in \{0,1\} \}$, that is, the constraint ${x} \in D$ is binding and the maximizer of $\im_n(\cdot)$ lies at its upper/lower bound. 
	A related phenomenon is the concentration of inputs in the top/left and bottom/right corners of $D$, which are associated with the highest uncertainty about the level set due to the confluence of the zero contour passing there and reduced spatial information from being on the edge of $D$.
	
	Another noteworthy feature is \emph{replication} of some inputs, that is, repeated selection of the same $\mathbf{x}$ sites. This does not occur for MCU, but happens for ICU, tMSE and cSUR that frequently (across algorithm runs) sample repeatedly at the vertices of $D$ (indicated by the size of the corresponding marker in Figure~\ref{2dexpresult}). The replication is typically mild (we observe 145+ unique designs among a total of 150 $x_n$'s). This finding echoes~\cite{binois2017replication} the importance for the metamodel to distinguish between signal and noise, which is a key distinction with the noise-free setting $\epsilon(x) \equiv 0$.
	
	 Given the above discussion and the relative overhead of the different heuristics, we conclude that in lower dimensional problems, there is little benefit to using the more sophisticated ICU criterion, while for higher dimensional problems, ICU criterion is significantly better than the others. Beyond that, tMSE appears to be  an adequate and cheaper choice. However, as the input space becomes more complicated, we need more exploration over the input space and the explorative criteria like ICU start to shine.

	The performance of designs differs when combined with different GP metamodels. Table~\ref{tbl:2d} shows that there is no one overall ``best" design for all metamodels across all cases. However, it does suggest some design/metamodel ``combos" that work better than others, especially in the 1-D and 2-D experiments. The classification GPs seem to prefer more aggressive designs, such as MCU, while the regression GPs prefer more exploratory designs, such as ICU. In higher dimensions, ICU usually wins across all metamodels in accuracy; see the results of 6-D experiments in Table~\ref{tbl:2d}.

\section{Application to Optimal Stopping Problems in Finance} \label{sec:Bermudan}

In our next case study we consider contour finding for determining the optimal exercise policy of a Bermudan financial derivative, cf.~Section~\ref{motivation}. The underlying simulator is based on a $d$-dimensional geometric Brownian motion $(\bm{X}_{t})$ that represents prices of $d$ assets and follows the log-normal dynamics
\begin{align}\label{eq:brown}
\bm{X}_{t+\Delta t}&= \bm{X}_{t} \exp \bigg((r-\frac{1}{2}\sigma^2)\Delta t +\bm{\Sigma} \Delta \bm{W}_{t}\bigg), \qquad
 \Delta \bm{W}_{t} &\sim \mathcal{N}(0,\Delta t \bm{I}),
\end{align}
where $\bm{I}$ is the $d\times d$ identity matrix.
Let $h(t,x)$ be the option payoff from exercising when $\bm{X}_{t} = x \in \mathbb{R}^d$. 
Exercising is allowed every $\Delta t$ time units, up to the option maturity $T$, so that we wish to determine the collection $\{ S_t : t \in \{ \Delta t, 2\Delta t, \ldots, T-\Delta t\} \}$, which are the zero level sets of the timing function $ x\mapsto T(t,x)$.  During the backward dynamic programming, we iterate over $t = T, T-\Delta t, \ldots, 0$,
and the simulator of $T(t,x)$ returns the difference between the pathwise payoff along a trajectory of $(\bm{X}_{t:T})$ that is based on the forward exercise strategy summarized by the forward-looking $\{ \hat{S}_s, s > t\}$, and $h(t,x)$.

As discussed in \cite{ludkovski2015kriging}, this setting implies a skewed, non-Gaussian, heteroskedastic distribution of the simulation noise and is a challenging stochastic contour-finding problem. Note that in order to reflect the underlying distribution of $\bm{X}_{t}$ at time $t$ (conditional on the given initial value $\bm{X}_0 = x_0$) the weighting measure $\mu(x)= p_{X_t}( x | x_0)$ is used. Thus, $\mu(\cdot)$ is log-normal based on \eqref{eq:brown} and is multiplied by the respective $\im_n$ criteria when selecting $x_{n+1}= \arg \max_{x \in D } \im_{n}(x) \mu(x)$.
In line with the problem context, we no longer directly measure the accuracy of learning $\{ S_{t} \}$ but instead focus on the ultimate output of RMC, which is the estimated option value in \eqref{payoff}. The latter must itself be numerically evaluated via an out-of-sample Monte Carlo simulation that averages realized payoffs along a large database of $M$ paths $x^{1:M}_{0:T}$:
\begin{align}\label{eq:hat-V}
\hat{V}(0, x_0) &= \frac{1}{M} \sum_{m=1}^M h(\tau^m, x^{(m)}_{\tau^m} ), \qquad 
 \tau^m = \inf \{t : x^{(m)}_{t} \in \hat{S}_{t} \}.
 \end{align}
Since our goal is to find the \emph{best} exercise value, higher $\hat{V}$'s indicate a better approximation of $\{S_t\}$.

To allow a direct comparison, we set parameters matching the test cases in \cite{ludkovski2015kriging}, considering a 2-D and 3-D example. In both cases the volatility matrix $\bm{\Sigma} = \sigma \bm{I}$ in \eqref{eq:brown} is diagonal with constant terms; that is, the coordinates
$\mathbf{X}_{1:n}^1,\ldots,\mathbf{X}_{1:n}^d$ are independently and identically distributed. As a first example, we consider a 2-D basket Put option with parameters $r=0.06, \sigma=0.2, \Delta t = 0.04, \cK=40, T= 1$. The payoff is $h(t,x) = e^{-rt}(\cK-\frac{x^1 + x^2}{2})_+$ with $K=40$. Here it is known that stopping becomes optimal once both asset prices $x^1$ and $x^2$ become sufficiently low, so the level set $S_{t}$ is toward the bottom-left of $D$; see Fig~\ref{fig:2doption}. In contrast, stopping is definitely suboptimal when $h(t,x) = 0 \Leftrightarrow (x^1+x^2)/2 > \cK$. Consequently, the input sample space is taken to be $D = [25,55] \times [25, 55] \cap \{x^1+x^2 \leq 80\}$.

In this first case study, the timing value $h(t, x)$ is known to be \emph{monotonically} increasing in the asset price $x$. To incorporate this constraint, we augment the four main metamodels (GP, $t$-GP, Cl-GP and TP) with two monotonic versions, M-GP and MCl-GP.  By constraining the fitted $\hat{f}$ to be monotone, we incorporate structural knowledge about the ground truth, which in turn reduces posterior uncertainty and thus might produce more accurate estimates of $S$. Monotonicity of the metamodel for $f$ is also one sufficient way to guarantee that the outputted level set $\hat{S}$ is a \emph{connected} subset of $D$.

	Our monotone GPs are based on~\cite{riihimaki2010gaussian}. In general, any infinite-dimensional Gaussian process is intrinsically non monotone, since the multivariate Gaussian distribution is always supported on the entire $\mathbb{R}^d$, rather than an orthant. Nevertheless, local monotonicity in $\hat{f}$ can be enforced by considering the gradient $\nabla f$ of $f$ which is also a Gaussian process. Specifically, \cite{riihimaki2010gaussian} proposed to adaptively add  virtual observations for $\nabla f$; we employ the resulting implementation in the public \texttt{GPstuff} library~\citep{vanhatalo2012bayesian} to build our own version dubbed M-GP. We employ the same strategy to restrict the coordinates $z^j$ of the latent logistic GP $Z$ to be increasing (decreasing) across $D$. Implementation details are included in Appendix~\ref{app:mgp}.

As a second example, we consider a 3-D max-Call $x \in \mathbb{R}^3$ with payoff $h(t,x) = e^{-rt}(\max(x^1, x^2, x^3) - \cK)_+$. The parameters are $r = 0.05, \delta = 0.1, \sigma = 0.2, X_0 = (90,90,90), \cK = 100, T = 3$ and $\Delta t = 1/3$. Since stopping is ruled out when $h(t,x) = 0 \Leftrightarrow \max(x^1, x^2, x^3) < \cK$, the sample space is taken to be $D = [50,150]^3 \cup \{\max(x^1, x^2, x^3) > \cK\}$. In this case, stopping is optimal if \emph{one} of the coordinates $x^i$ is significantly higher than the other two, so $S_{t}$ consists of three disconnected components. In this problem there is no monotonicity, so we employ only the GP, t-GP, Cl-GP, and TP metamodels.

Because of the iterative construction of the simulator, the signal-to-noise ratio gets low for small $t$'s. The variance $\taun^2(x)$ is also highly state-dependent, tending to be smaller for sites further from the zero-contour.  
To alleviate this misspecification and reduce metamodel overhead, we employ \emph{batched} designs \citep{ludkovski2015kriging,ankenman2010stochastic}, reusing $x \in D$ for $r$ replications to collect  observations $y^{(1)}(x),\ldots,y^{(r)}(x)$ from the corresponding simulator $Y(x)$. Then, we treat the mean of the $r$ observations,
\begin{align}
\bar{y}(x) = \frac{1}{r}\sum_{i=1}^r y^{(i)}(x),
\end{align}
as the response for input $x$ and use $(x,\bar{y}(x))$ as a single design entry. The statistical properties of $\bar{y}$ are improved compared with the raw observations $y$: it is more consistent with the Gaussian assumption thanks to the Central Limit Theorem (CLT), and its noise variance $\bar{\taun}^2(x) = \taun^2(x)/r$ is much smaller. Since the expense of sequential design of GP metamodels comes mainly from choosing the new input at each step, the reduction in budget $n=N/r$ by a factor of $r$ significantly speeds their fitting and updating, with $n$ for the number of unique inputs.

For the 2-D Put case study, we then test a total of three budget settings: (i) $r=3, n = 80$ (low budget of $N=240$ simulations); (ii) $r=15, n = 80$ (high budget $N=800$ with moderate replication); (iii) $r=48, n = 25$ (high $N=800$ with high replication). Comparing (ii) and (iii) shows the competing effects of having non-Gaussian noise (for lower $r$) and small design size (low $n$). The initial design size  $n_0=10$.  In this example, taking $n \gg 80$ gives only marginally better performance but significantly raises the computation time and hence is ruled out as impractical. Three setups are investigated for the 3-D example: $r = 3, n= 100$ (low-budget of $N=300$), $r = 20, n = 100$ (moderate-budget of $N=2000$) and $r = 20, n = 200$ (high budget  $N=4000$), both with $n_0 = 30$. In all examples, the results are based on 25 runs of each scheme and are evaluated through the resulting expected reward $\hat{V}(0,x_0)$ \eqref{eq:hat-V} on a fixed out-of-sample testing set of $M=160,000$ paths of $\bm{X}_{0:T}$.

\begin{table*}[htb]
	\caption{Performance of different designs and models on the 2-D Bermudan Put option in Section~\ref{sec:Bermudan}. Results are the mean (standard deviation) payoff of 25 runs of experiments evaluating on the same out-of-sample testing set of $M=160000$ $\bm{X}_{0:T}$-paths at each run.}
		\centering
		\begin{tabular}{llllll}
			\hline\noalign{\smallskip}
			& LHS  & MCU & tMSE &cSUR & ICU \\
			\noalign{\smallskip}\hline\noalign{\smallskip}
			\multicolumn{6}{c}{$\mathbf{R = 3, n^* = 80}$} \\
			\noalign{\smallskip}\hline\noalign{\smallskip}
			GP &  1.211(0.120) &   1.425(0.008)  &  1.427(0.007)  &  1.431(0.009)  &  1.431(0.007)\\
			$t$-GP & 1.125(0.113)  &  1.409(0.013)  &  1.417(0.008)  &  1.409(0.010)  &  1.406(0.013) \\
			TP &  1.179 (0.133)  & 1.408 (0.022)  & 1.414 (0.008) &  1.378 (0.044) & 1.316 (0.037) \\			
			M-GP & 1.403(0.014)   & 1.438(0.007)  &  1.440(0.006)   & 1.442(0.009)   & 1.433(0.005) \\
			Cl-GP & 1.111(0.121)   & 1.395(0.015)   & 1.402 (0.013)  & 1.393(0.013)   & 1.391(0.013) \\
			MCl-GP & 1.407(0.008)   & 1.429(0.010)   & 1.429(0.013)   & 1.431(0.007)   & 1.396(0.019) \\
			\noalign{\smallskip}\hline\noalign{\smallskip}
			\multicolumn{6}{c}{$\mathbf{R = 15, n^* = 80}$} \\
			\noalign{\smallskip}\hline\noalign{\smallskip}
			GP  & 1.425 (0.017) & 1.448 (0.003)  &  1.450 (0.002)  &  1.450 (0.003) & 1.449 (0.003)\\
			$t$-GP & 1.406 (0.033) &  1.445 (0.003)  &  1.447 (0.002)  &  1.444 (0.005) & 1.446 (0.004)\\
			TP &  1.414 (0.023)   & 1.443 (0.003)  & 1.443 (0.004) & 1.441 (0.004) & 1.430 (0.006) \\
			M-GP & 1.407 (0.008)  &  1.449 (0.003)  &  1.451 (0.002) & 1.454 (0.002) & 1.451 (0.003)\\
			Cl-GP& 1.353 (0.050)  &  1.441 (0.004) & 1.440 (0.003)  &  1.435 (0.004) & 1.436 (0.005)\\
			MCl-GP& 1.416 (0.010)  &  1.448 (0.004)  &  1.449 (0.003)  &  1.443 (0.003) & 1.418 (0.008)\\
			\noalign{\smallskip}\hline\noalign{\smallskip}
			\multicolumn{6}{c}{$\mathbf{R = 48, n^* = 25}$} \\
			\noalign{\smallskip}\hline\noalign{\smallskip}
			GP  &  1.341 (0.068) &  1.450 (0.003) &  1.449 (0.003) &  1.443 (0.004) & 1.448 (0.003)\\
			$t$-GP & 1.336 (0.126) &  1.449 (0.003) &  1.452 (0.003)  &  1.442 (0.004) & 1.449 (0.003)\\
			TP &  1.367 (0.063)  & 1.433 (0.006) & 1.430 (0.011) & 1.421 (0.039) &  1.423 (0.023) \\
			M-GP & 1.415 (0.007)  & 1.446 (0.002)  &  1.444 (0.002)  &  1.445 (0.004) & 1.442 (0.004) \\
			Cl-GP& 1.110 (0.144) & 1.430 (0.010)  &  1.434 (0.005)  &  1.409 (0.008)  &  1.388 (0.016) \\
			MCl-GP& 1.423 (0.015) & 1.446 (0.004) & 1.448 (0.003)  &  1.413 (0.024)  &  1.414 (0.024) \\
			\noalign{\smallskip}\hline
		\end{tabular}
	\label{tbl:2doption}
\end{table*}

\subsection{Results}

Tables \ref{tbl:2doption} and \ref{tbl:3doption} compare the different designs and metamodels. To assess the sequential design gains, we also report the results from using a baseline nonadaptive LHS design on $D$.  At low budget, we observe the dramatic gains of using adaptive designs for level set estimation, which allow us to obtain the same performance with an order-of-magnitude smaller simulation budget. The tMSE and cSUR criteria work  best for the 2-D Put, while ICU is the best for the 3-D max-Call, indicating that the exploratory designs start to win out in more complex settings with higher $d$.

Regarding the metamodels, in the low-budget setups, the monotonic GP metamodel works best for the 2-D Put and $t$-GP for the 3-D max-Call. For the higher budget, which also coincides with higher $r \in \{10, 50\}$, the metamodel performance is similar, with $t$-GP slightly better than the other GP variants. In particular, once the SNR is high, classical Gaussian GP is effectively as good as any alternative. In both examples, TP and classification metamodels do not work well, possibly because of being more sensitive to the heteroscedastic aspect. We note that TP as well as the classification metamodels suffer from instability, so that lower $\hat{V}(0,x_0)$ is matched with a high sampling standard deviation. Another observation is that Cl-GP and MCl-GP perform badly with exploratory heuristic like ICU, especially with high budget.

\begin{figure*}[htb]
	\begin{minipage}[t]{0.33\linewidth}
		\begin{center}
			\includegraphics[width=1\textwidth,trim=0.3in 0.2in 0.3in 0.2in]{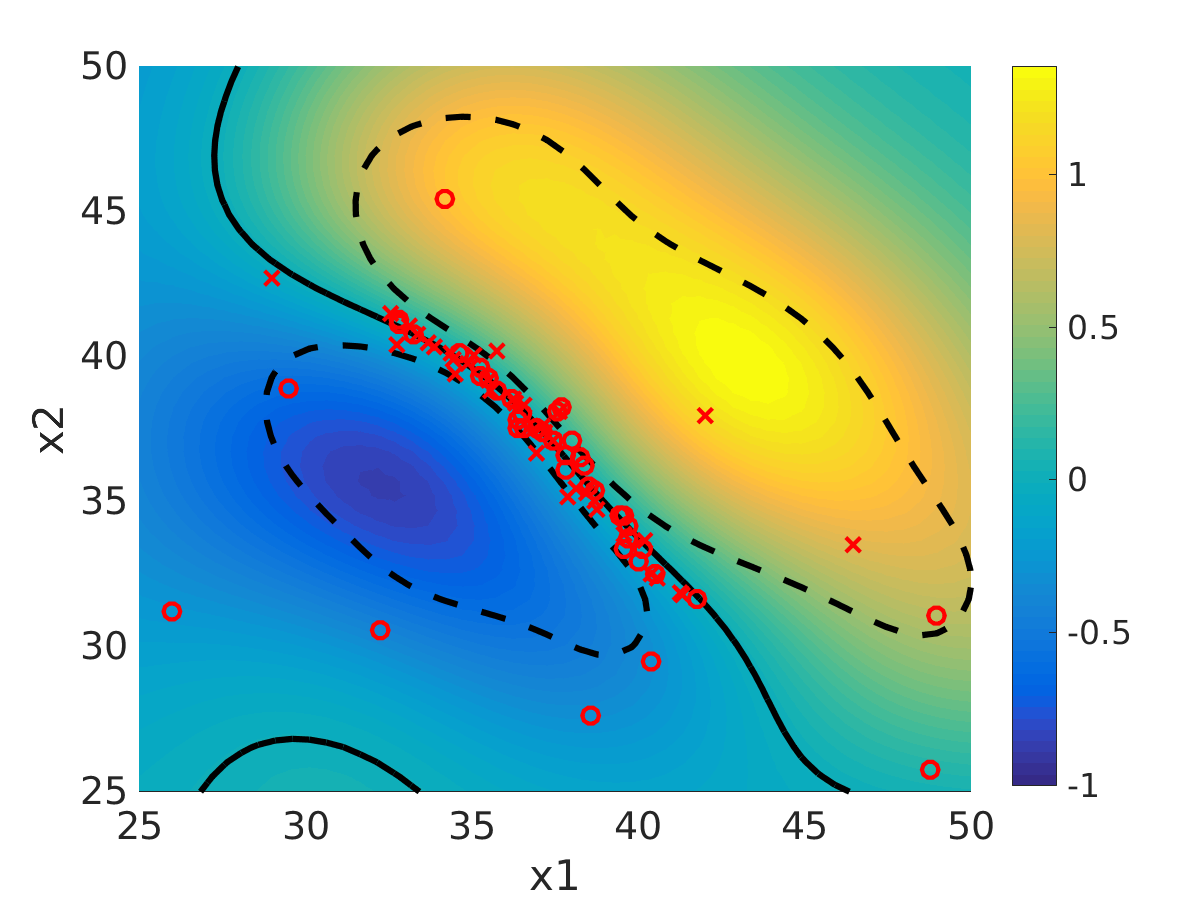} \\
			(a) GP with tMSE 
		\end{center}
	\end{minipage}
	\begin{minipage}[t]{0.33\linewidth}
		\begin{center}
			\includegraphics[width=1\textwidth,trim=0.3in 0.2in 0.3in 0.2in]{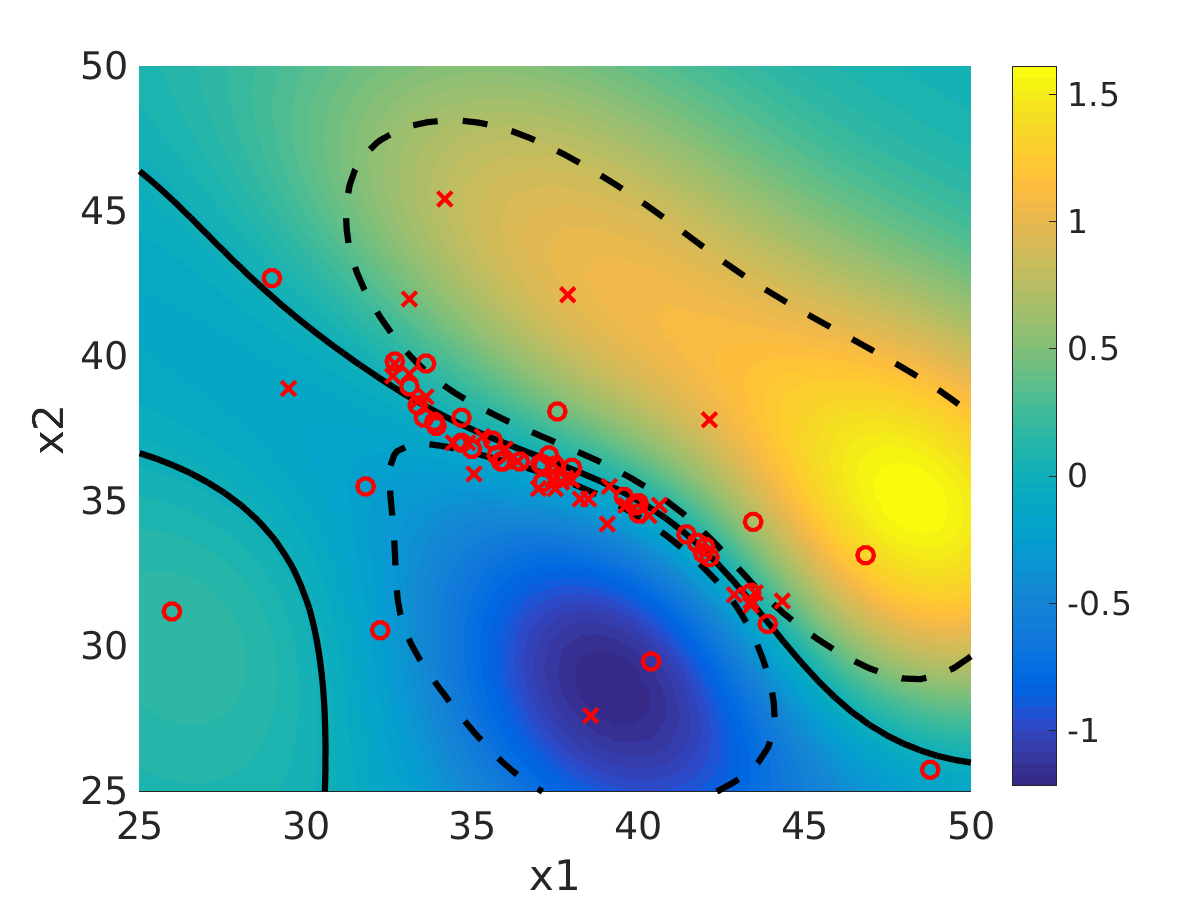} \\
			(b) $t$-GP with tMSE
		\end{center}
	\end{minipage}
	\begin{minipage}[t]{0.33\linewidth}
		\begin{center}
			\includegraphics[width=1\textwidth,trim=0.3in 0.2in 0.3in 0.2in]{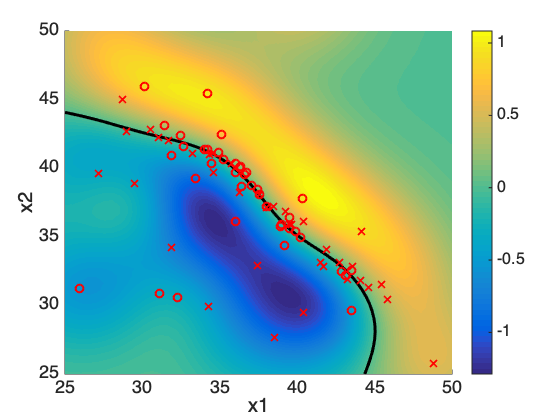} \\
			(c) Cl-GP with MCU
		\end{center}
	\end{minipage}
	\begin{minipage}[t]{0.33\linewidth}
		\begin{center}
			\includegraphics[width=1\textwidth,trim=0.3in 0.2in 0.3in 0in]{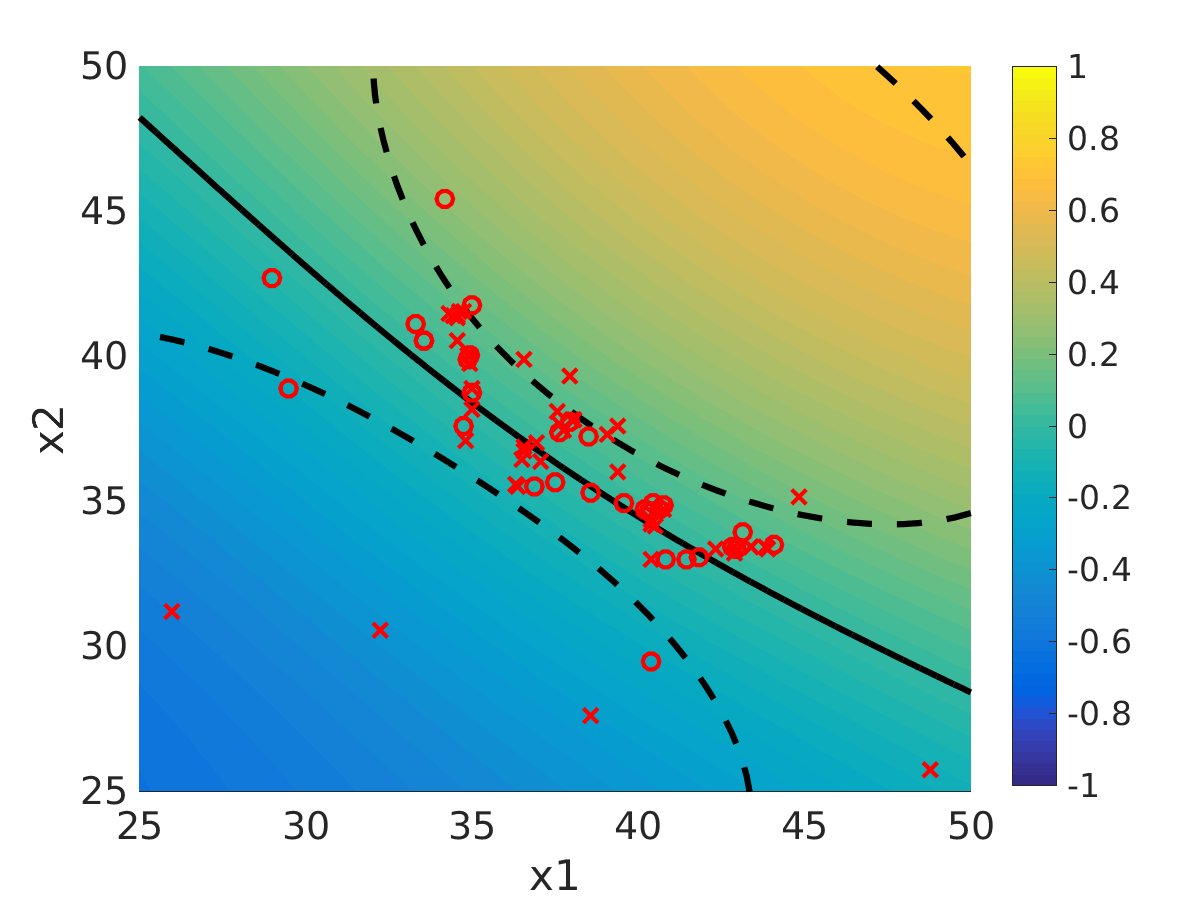} \\
			(d) M-GP with cSUR
		\end{center}
	\end{minipage}
	\begin{minipage}[t]{0.33\linewidth}
		\begin{center}
			\includegraphics[width=1\textwidth,trim=0.3in 0.2in 0.3in 0in]{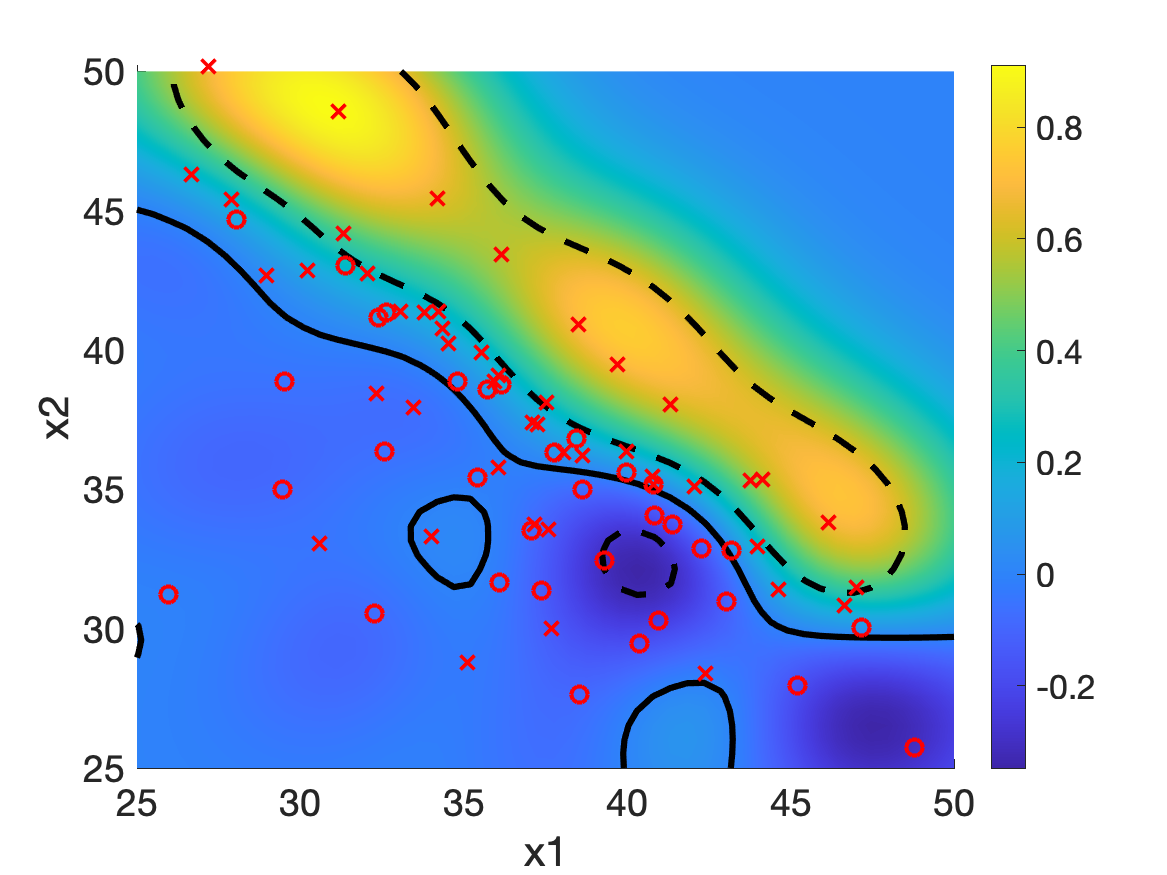} \\
			(e) TP with cSUR
		\end{center}
	\end{minipage}
	\begin{minipage}[t]{0.33\linewidth}
		\begin{center}
			\includegraphics[width=1\textwidth,trim=0.3in 0.2in 0.3in 0in]{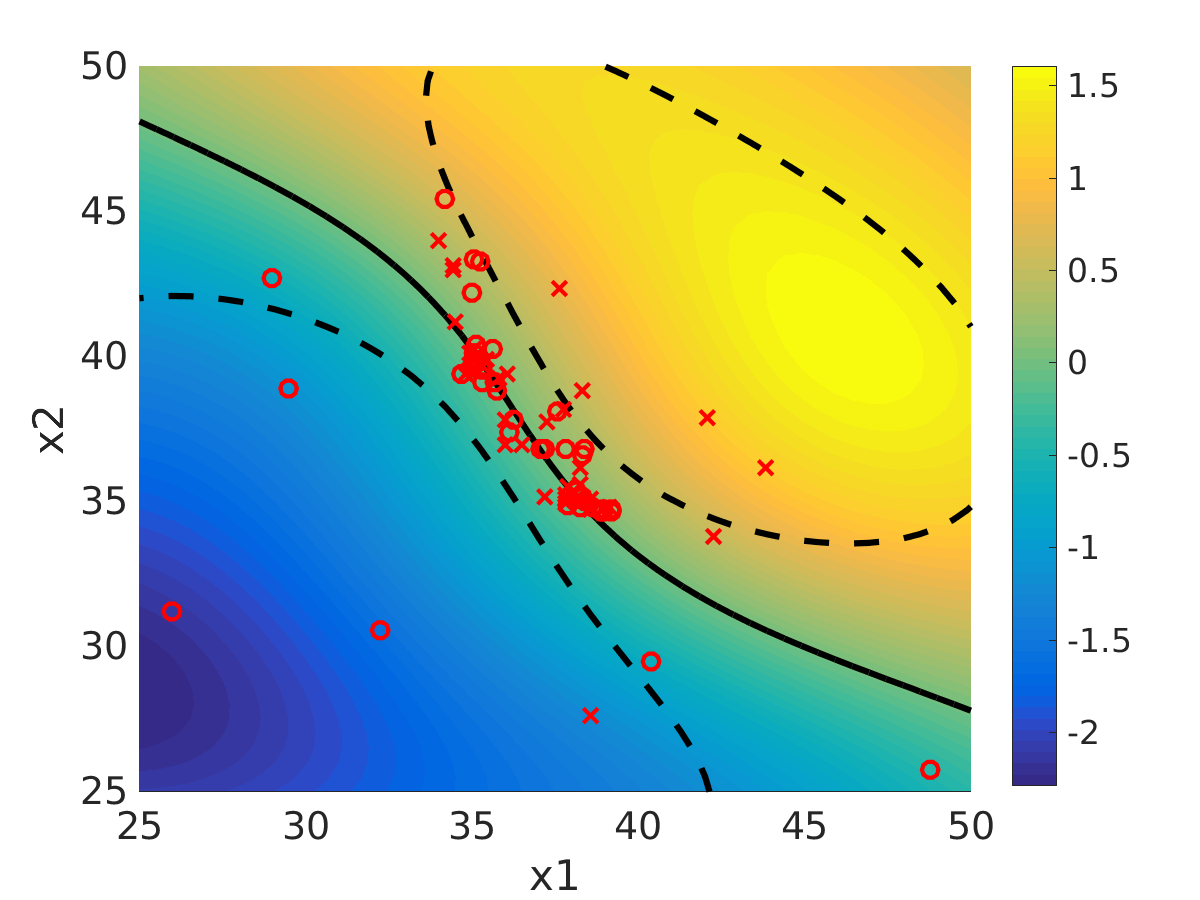} \\
			(f) MCl-GP with tMSE
		\end{center}
	\end{minipage}
	\caption{The estimated exercise boundary $\partial \hat{S}$ (solid line with 95\% CI as dashed lines) at $t = 0.6$ for 2-D Bermudan Put from Section~\ref{sec:Bermudan}. Shading, which varies panel to panel, indicates the point estimate for the latent  $\hat{f}(x)$ or $\hat{z}(x)$. We also show the design $(\bx,\by)$ with positive $y_n$'s marked by $\times$ and negative $y_n$'s by $\circ$.  All schemes used  $R = 15, n^* = 80$. }
	\label{fig:2doption}
\end{figure*}

Figure \ref{fig:2doption} shows the estimated exercise boundary  $\partial \hat{S}_{t}$ with its 95\% CI at $t = 0.4$ for the 2-D Put, for each of the five metamodels, each with the design yielding the highest payoff. We observe that all the best-performing designs look similar, placing about a dozen $x_n$'s (some of which are from the initial design $x_{1:n_0}$) throughout $D$ and the rest tightly along the zero contour. The results suggest that the criteria are largely interchangeable and that simpler $\im_n$ heuristics are able to reproduce the features of the more sophisticated or expensive ICU. The heuristics \emph{do} differ in their uncertainty quantification; $t$-GP and GP generate tightest CI bands, while those of classification GPs and TP are too wide, indicating lack of confidence in the estimate. Of note, the regression GP metamodels (GP, $t$-GP and M-GP) also generate the lowest sampling variance for $\hat{V}(0,x_0)$.

Based on these results, our take-aways are threefold. First, similar to \cite{ludkovski2015kriging} we document significant gains from sequential design.Second, we find that while using ICU is helpful for more complicated settings with higher dimension $d$ and larger budget, tMSE is the recommended DoE heuristic for lower dimensional cases, achieving excellent results with minimal overhead (in particular without requiring look-ahead variance). Third, we find that for applications with thousands of simulations, the Gaussian observation model is sufficient, since the underlying design needs to be replicated $r \gg 1$ in order to avoid excessively large $\mathbf{K}$-matrices. Therefore, there is little need for more sophisticated metamodels, although useful gains can be realized from enforcing the monotonic structure, if available.

\begin{table*}[htb]
	\caption{Performance of different designs and models on the 3-D Bermudan max-Call in Section~\ref{sec:Bermudan}. Results are the mean (w/standard deviation) payoff of 25 macroreplications evaluating on the same out-of-sample testing set of $M=160000$ $\bm{X}_{0:T}$-paths at each run.}
		\centering
		\begin{tabular}{lrrrrr}
			\hline\noalign{\smallskip}
			& LHS  & MCU & tMSE &cSUR & ICU \\
			\noalign{\smallskip}\hline\noalign{\smallskip}
			\multicolumn{6}{c}{$\mathbf{R = 3, n^* = 100}$} \\
			\noalign{\smallskip}\hline\noalign{\smallskip}
			GP  & 10.036 (0.331) & 10.725 (0.095)  & 10.773 (0.071) &  10.711 (0.086)  & 10.753 (0.072) \\
			$t$-GP & 9.894 (0.447)  & 10.736 (0.088)  & 10.747 (0.087)  & 10.720 (0.104)  & 10.782 (0.076) \\
			TP & 9.169 (0.354)  & 10.101 (0.218)  & 9.872 (0.102)  & 8.867 (0.357)  & 10.482 (0.156) \\ 			
			Cl-GP & 9.552 (0.567)  & 10.566 (0.084)  & 10.657 (0.097)  & 10.586 (0.099) &  10.604 (0.119) \\
			\noalign{\smallskip}\hline\noalign{\smallskip}
			\multicolumn{6}{c}{$\mathbf{R = 20, n^* = 100}$} \\
			\noalign{\smallskip}\hline\noalign{\smallskip}
			GP & 10.924 (0.076) & 11.078 (0.029) &   11.072 (0.028) &   11.055 (0.032) &  11.101 (0.023) \\
			$t$-GP & 10.923 (0.071) &   11.061 (0.039) &  11.055 (0.027) &   11.044 (0.029) &   11.100 (0.027) \\
			TP & 10.385 (0.178) & 10.815 (0.039)  & 10.745 (0.045)  & 10.620 (0.087)  & 10.507 (0.087)   \\
			Cl-GP & 10.761 (0.112) &   11.026 (0.032) &   10.991 (0.037) &   10.901 (0.049) &   10.937 (0.041) \\
			\noalign{\smallskip}\hline\noalign{\smallskip}
			\multicolumn{6}{c}{$\mathbf{R = 20, n^* = 200}$} \\
			\noalign{\smallskip}\hline\noalign{\smallskip}
			GP & 11.105(0.036) & 11.147(0.021) &   11.119(0.022) &   11.131(0.018) &  11.178(0.020) \\
			$t$-GP & 11.090(0.034) &   11.141(0.019) &  11.126(0.020) &   11.115(0.027) &   11.175(0.021) \\
			TP & 10.585 (0.118) & 10.896 (0.030) & 10.811 (0.035) & 10.764 (0.041) & 10.638 (0.038) \\
			Cl-GP & 10.995(0.059) &   11.109(0.025) &   11.056(0.040) &   10.985(0.027) &   11.010(0.029) \\
		\noalign{\smallskip}\hline
		\end{tabular}
	\label{tbl:3doption}
\end{table*}

\section{Conclusion}\label{sec:conc}

We have carried a comprehensive comparison of five metamodels and four design heuristics on 18 case studies ($4 \times 3$ synthetic, plus six real-world). In sum, the considered alternatives to standard Gaussian-observation GP do perform somewhat better. In particular, $t$-GP directly nests plain GP and hence essentially always matches or exceeds the performance of the latter. We also observe gains from using Cl-GP when SNR is low and from monotonic surrogates when the underlying response is monotone. That being said, final recommendation regarding the associated benefit depends on computational considerations, as the respective overhead becomes larger (and exact updating of the metamodel no longer possible).

In terms of design, we advocate the benefits of tMSE in low dimensional simulations, which generates high-performing experimental designs without requiring expensive acquisition function (or even look-ahead variance). The tMSE criterion does sometimes suffer from the tendency to put many designs at the edge of the input space but otherwise tends to match the performance of more complex and computationally intensive $\im_n$'s. For complex simulations, ICU is probably still the best choice (although in that case, random-set-based heuristics should also be considered). Especially in higher dimensions with misspecified noise, ICU is the best choice among all designs. We also stress that the user ought to thoughtfully pick the \emph{combination} of sequential design and metamodel, since cross-dependencies are involved  (e.g., classification metamodels generally do not work well with the ICU criterion in lower dimension).

\subsection*{Acknowledgements}
XL and ML are partially supported by NSF DMS-1521743 and DMS-1821240. The work of MB is partially supported by NSF DMS-1521702 and the U.S.~Department of Energy, Office of Science, Office of Advanced Scientific Computing Research under Contract No.\ DE-AC02-06CH11357.

\bibliographystyle{spbasic}      

\bibliography{jobnames}   

\appendix

\section{Choice of $\gamma^{(n)}$ for MCU} \label{app:mcu}

\begin{figure*}[htb]
	\begin{minipage}[t]{0.33\linewidth}
		\begin{center}
			\includegraphics[width=1\textwidth,trim=0.3in 0.2in 0.3in 0.2in]{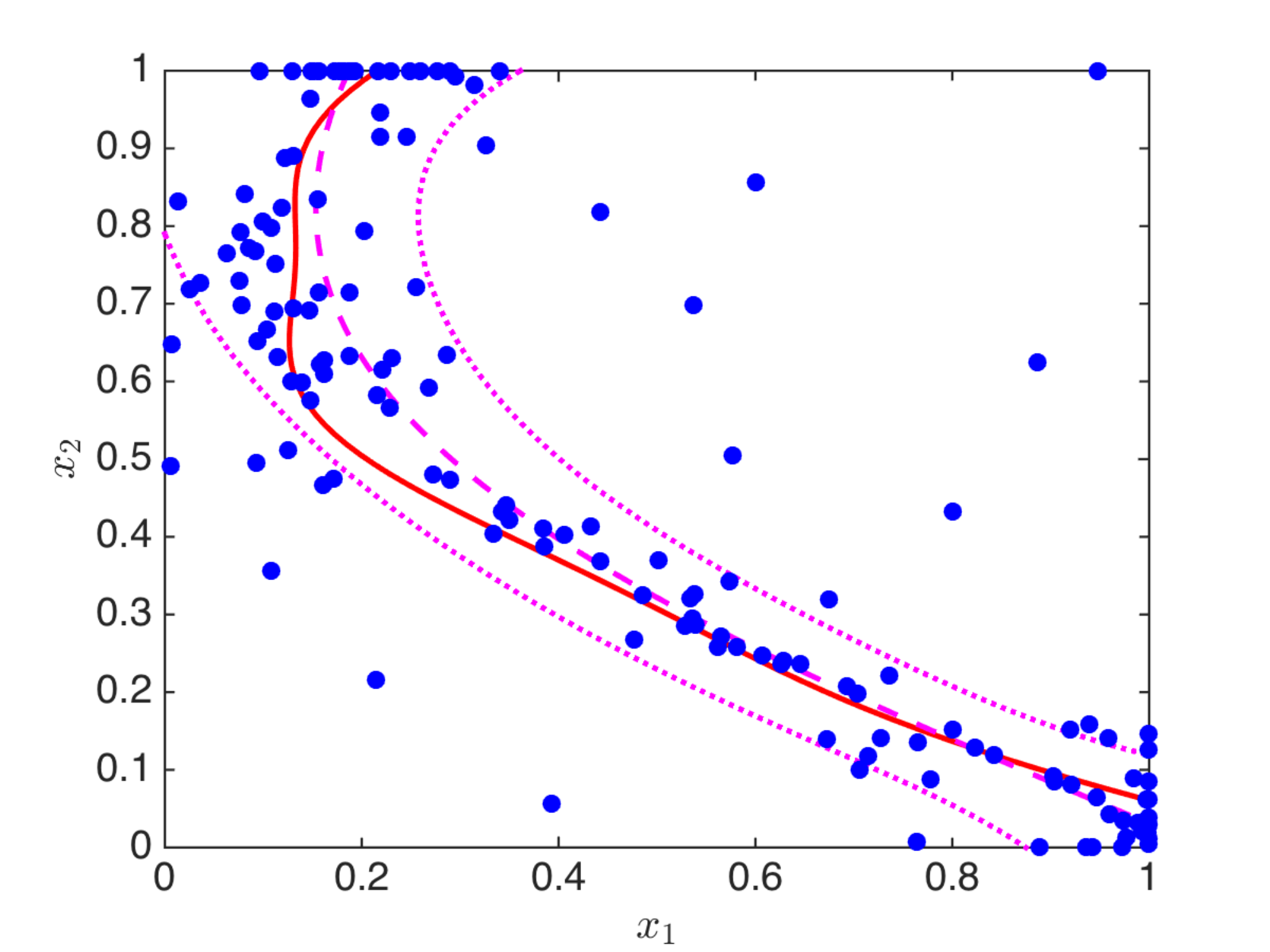} \\
			(a) $\gamma^{(n)} = 0.5$
		\end{center}
	\end{minipage}
	\begin{minipage}[t]{0.33\linewidth}
		\begin{center}
			\includegraphics[width=1\textwidth,trim=0.3in 0.2in 0.3in 0.2in]{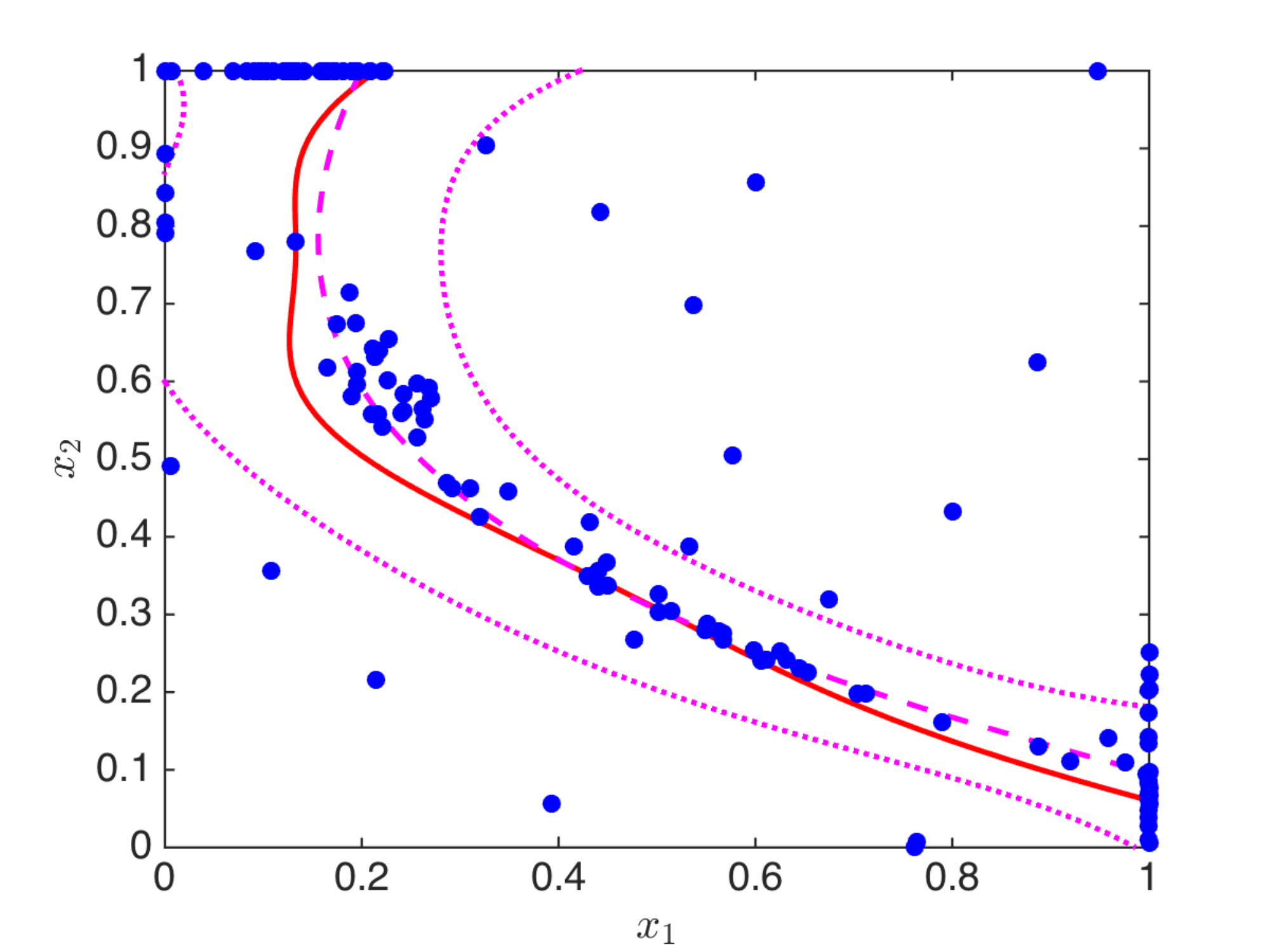} \\
			(b) $\gamma^{(n)} = 1.96$
		\end{center}
	\end{minipage}
	\begin{minipage}[t]{0.33\linewidth}
		\begin{center}
			\includegraphics[width=1\textwidth,trim=0.3in 0.2in 0.3in 0.2in]{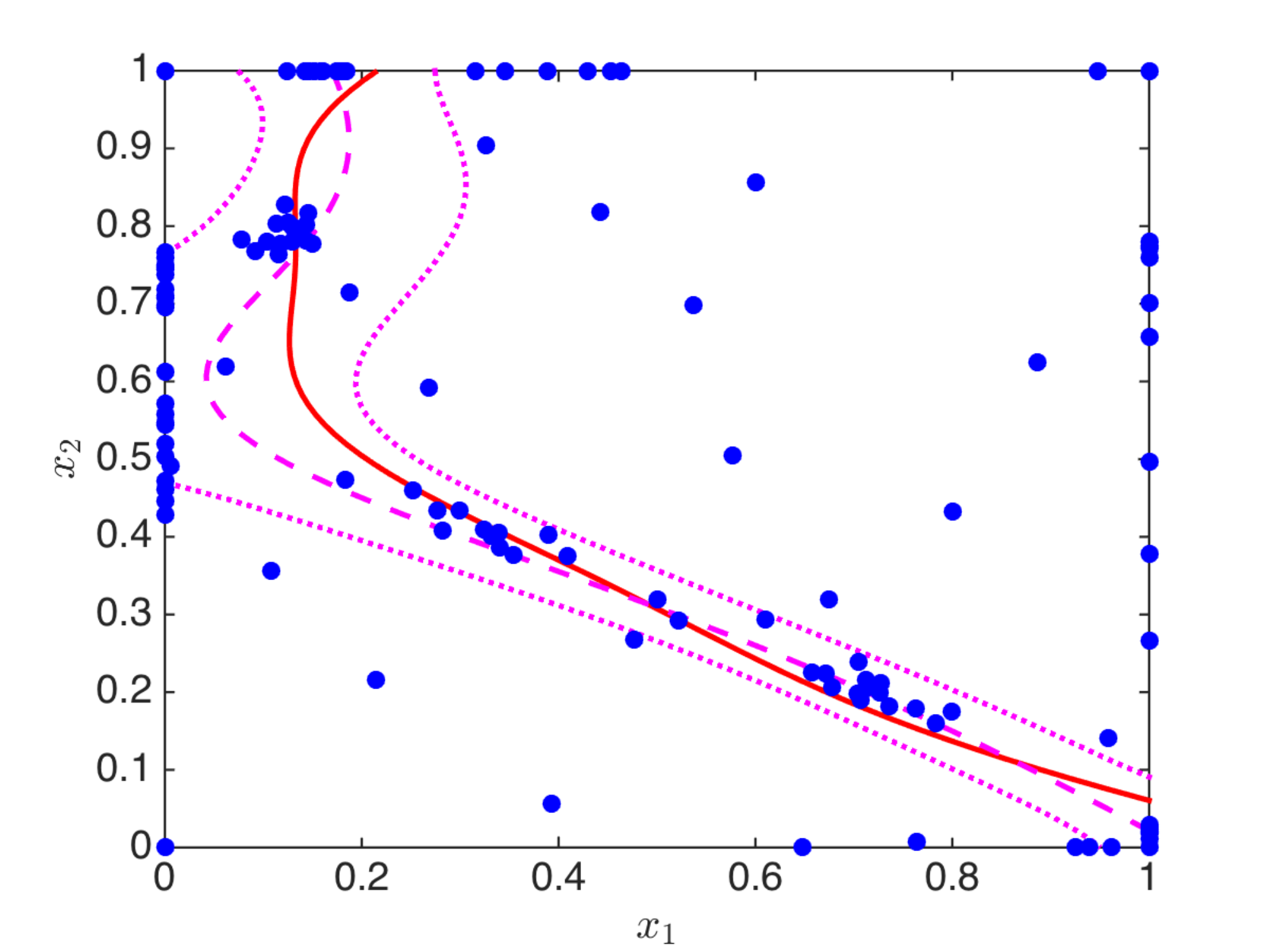} \\
			(c) $\gamma^{(n)} = 10$
		\end{center}
	\end{minipage}
	\caption{The estimated boundary $\partial \hat{S}$ (dashed line with 95\% CI as dotted lines). Blue dots are samples selected by MCU with $\gamma^{(n)} = 0.5$, $\gamma^{(n)} = 1.96$, and $\gamma^{(n)} = 10$.}
	\label{fig:2d-gamma}
\end{figure*}

Following up on the discussion in Section~\ref{sec:improvementmetrics}, we investigate the role of $\gamma^{(n)}$ in the performance of MCU. Table~\ref{tbl:MCU-gamma} shows the error rate $\cR$ for GP and $t$-GP metamodels with MCU acquisition function in the 2D synthetic experiments across three constant values of $\gamma^{(n)}$ (constant $\gamma^{(n)}$ was also employed in~\cite{gotovos2013active} and~\cite{bryan2006active}). We observe that generally the impact of $\gamma^{(n)}$ is secondary (with Gaussian GP being more sensitive), and moreover there is no single choice that works the best across all cases. As illustrated in Figure \ref{fig:2d-gamma}, large $\gamma^{(n)}$ favors space-filling, while small $\gamma^{(n)}$ favors exploitation in regions close to the boundary $\partial S$.
	
	Generally, smaller $\gamma^{(n)}$'s work better in cases with less noise (e.g.~$\gamma^{(n)} = 10$ is worst in the $t/ \text{small}$ noise scenario). This validates our recommendation that $\gamma^{(n)}$ should be adaptive to the signal-to-noise ratio of $\hat{f}^{(n)}(x)$ and $s^{(n)}(x)$. Clearly $s^{(n)}(x)$ depends strongly on the original noise specification which is another reason why a fixed ``universal'' $\gamma^{(n)}$ is inappropriate (unlike in deterministic experiments, there is no simple way to normalize the variance of $\epsilon$). Note that since $s^{(n)}$ decreases in $n$, the signal-to-noise ratio increases over time, which is consistent with the theoretical results that $\gamma^{(n)}$ should increase with $n$.

\begin{table}[htb]
	\centering
	{\small 		\begin{tabular}{lllllllllllll}
			\hline\noalign{\smallskip}
			Model &  $\gamma^{(n)} = 0.5$  & $\gamma^{(n)} = 1.96$   & $\gamma^{(n)} = 10$ \\
			\noalign{\smallskip}\hline\noalign{\smallskip}
			\multicolumn{4}{c}{$\mathbf{t/ \text{small}}$} \\
			\noalign{\smallskip}\hline\noalign{\smallskip}
			GP   &  1.87\% (0.36\%) &    1.82\% ( 0.51\%) &    2.09\% (0.54\%)  \\
			$t$-GP  &    1.80\% (0.52\%) &    1.73\% (0.22\%) &    1.84\% (0.42\%) \\
			\noalign{\smallskip}\hline\noalign{\smallskip}
			\multicolumn{4}{c}{$\mathbf{t/ \text{large}}$} \\
			\noalign{\smallskip}\hline\noalign{\smallskip}
			GP    & 5.20\% (2.33\%) &    5.59\% ( 2.22\%) &   4.94\% (1.79\%)   \\
			$t$-GP  & 3.80\% (1.25\%) &    4.24\% (2.12\%) &    4.01\% ( 1.43\%)  \\
			\noalign{\smallskip}\hline\noalign{\smallskip}
			\multicolumn{4}{c}{$\mathbf{Gsn/ \text{mix}}$} \\
			\noalign{\smallskip}\hline\noalign{\smallskip}
			GP    & 5.10\% ( 2.36\%) &    5.53\% ( 1.79\%) &    6.01\% ( 3.08\%)  \\
			$t$-GP  & 4.63\% (1.74\%) &    3.92\% (1.26\%) &    4.39\% ( 1.40\%)  \\
			\noalign{\smallskip}\hline\noalign{\smallskip}
			\multicolumn{4}{c}{$\mathbf{t/ \text{hetero}}$} \\
			\noalign{\smallskip}\hline\noalign{\smallskip}
			GP &  11.23\% ( 5.08\%) &    10.52 \% (7.05\%) &    13.63\% (6.32\%)   \\
			$t$-GP &  7.34\% (3.96\%) &    10.58\% ( 8.25\%) &    7.77\% ( 3.55\%) \\
			\noalign{\smallskip}\hline
	\end{tabular}}
	\caption{Mean (w/standard deviation) error rate $\cR$ for MCU in 2D synthetic experiments. Results are based on 20 macro-replications of each scheme.}
	\label{tbl:MCU-gamma}
\end{table}

\section{Gaussian Process Classification} \label{app:clgp}

Similar to the $t$-GP, we use a Laplace approximation for the non-Gaussian $p_{\mathrm{Cl}}(\mathbf{\z}|\bx,\cby)$ in Eq.~(\ref{posteriorz}), 
\begin{align}
p_{\mathrm{Cl}}(\mathbf{\z}|\bx,\cby) &\approx  q_{\mathrm{Cl}}(\mathbf{\z}|\bx,\cby) = \mathcal{N}(\tilde{\mathbf{\z}}^{(n)},\mathbf{\Sigma}^{-1}_{\mathrm{Cl}}) , \label{cls}
\end{align}
where we again use the mode $\tilde{\mathbf{\z}}^{(n)}:=\arg \max_\mathbf{\z}p(\mathbf{\z}|\cA_n)$ and $\mathbf{\Sigma_{\mathrm{Cl}}}$ is the Hessian of the negative log posterior at $\tilde{\mathbf{\z}}^{(n)}$:
\begin{align}
\mathbf{\Sigma_{\mathrm{Cl}}}=-\nabla^2 \log p_{\mathrm{Cl}}(\mathbf{\z}|\cA_n)\big|_{\mathbf{\z}=\tilde{\mathbf{\z}}^{(n)}}=\mathbf{K}^{-1} +\mathbf{V},
\end{align}
and $\mathbf{V}=-\nabla^2 \log p(\cby|\mathbf{\z})|_{\mathbf{\z}=\tilde{\mathbf{\z}}^{(n)}}$ is diagonal with elements
\begin{align}\label{v}
v_i &= \mathbf{V}_{ii}=-\frac{\partial^2}{\partial \z_i^2} \log p(\check{y}_i|\z_i)\big|_{\z_i=\tilde{\z}^{(n)}_i} 
= \frac{\phi( \tilde{\z}^{(n)}_i)^2}{\Phi(\check{y}_i \tilde{\z}^{(n)}_i)^2} +  \frac{\check{y}_i \tilde{\z}^{(n)}_i\phi(\tilde{\z}^{(n)}_i)}{\Phi(\check{y}_i \tilde{\z}^{(n)}_i)},
\end{align}
for $i=1,\ldots,n$, $\phi(\cdot)$ denoting the density of the standard normal distribution.

\section{Computation Details for Look-Ahead Variance} \label{app:lookahead}

\textbf{$t$-GP:}
To approximate $\tilde{\mathbf{f}}_{t\mathrm{GP}}^{(n+1)}$ in $t$-GP, we recall that  the posterior mode and the posterior mean coincide:
\begin{align}
\hat{\mathbf{f}}^{(n)}_{t\mathrm{GP}}(\bx) = \mathbf{K} \mathbf{K}^{-1}\tilde{\mathbf{f}}^{(n)}_{t\mathrm{GP}}= \tilde{\mathbf{f}}^{(n)}_{t\mathrm{GP}}.
\end{align}
Hence we can compute the expected value of $\tilde{\mathbf{f}}^{(n+1)}_{t\mathrm{GP}}$ using the tower property:
\begin{equation}\label{towertmean}
\begin{split}
  \mathbb{E}[\tilde{\mathbf{f}}^{(n+1)}_{t\mathrm{GP}}|\bx,\by] 
 & = \mathbb{E}[\hat{\mathbf{f}}^{(n+1)}_{t\mathrm{GP}}(\mathbf{x}_{1:n+1})|\bx,\by] \\
 =& \mathbb{E} \left[\mathbb{E}[{f}(\mathbf{x}_{1:n+1})|\mathbf{x}_{1:n+1},\mathbf{y}_{1:n+1}]|\bx,\by \right] \\
 = & \mathbb{E}[{f}(\mathbf{x}_{1:n+1})|\bx,\by]  
 =  [\hat{\mathbf{f}}_{t\mathrm{GP}}^{(n)}(\mathbf{x}_{1:n}), \hat{f}_{t\mathrm{GP}}^{(n)}(x_{n+1})]
=  [\tilde{\mathbf{f}}_{t\mathrm{GP}}^{(n)}, \hat{f}_{t\mathrm{GP}}^{(n)}(x_{n+1})],
\end{split}
\end{equation}
where the last equality follows from the BLUP property of GP estimates. Therefore, we approximate the $(n+1)$-dimensional vector $\tilde{\mathbf{f}}^{(n+1)}_{t\mathrm{GP}}$ with $\check{\mathbf{f}}^{(n+1)}_{t\mathrm{GP}} =  [\tilde{\mathbf{f}}_{t\mathrm{GP}}^{(n)}, \hat{f}_{t\mathrm{GP}}^{(n)}(x_{n+1})]$, where the first component is $n$-dimensional and the second component is a scalar. In turn, this step allows us to update the matrices $\bm{W}^{(n)}_{t\mathrm{GP}}$ and $\bm{K}^{(n)}$ assuming a new input $x_{n+1}$ is added. Specifically, the new entry in $\bm{W}^{(n+1)}_{t\mathrm{GP}}$ is 
\begin{align}
\nonumber w^{(n+1)}_{n+1}  &= (\nu+1)\frac{\nu\taun^2-(y_{n+1}-\tilde{f}^{(n+1)}_{t\mathrm{GP}}(x_{n+1}))^2}{\left((y_{n+1}-\tilde{f}^{(n+1)}_{t\mathrm{GP}}(x_{n+1}))^2+\nu\taun^2 \right)^2} \\
&\simeq (\nu+1)\frac{\nu\taun^2-(y_{n+1}-\hat{f}^{(n)}_{t\mathrm{GP}}(x_{n+1}))^2}{\left((y_{n+1}-\hat{f}^{(n)}_{t\mathrm{GP}}(x_{n+1}))^2+\nu\taun^2 \right)^2}. \label{w}
\end{align}
Matching terms with the Gaussian observation GP,
the updated variance ${s}^{(n+1)}_{t\mathrm{GP}}(x_{n+1})^2$ is then approximately proportional to the current variance:
\begin{align}
\frac{{s}^{(n+1)}_{t\mathrm{GP}}(x_{n+1})^2}{s^{(n)}_{t\mathrm{GP}}(x_{n+1})^2} &\simeq  \frac{(w^{(n+1)}_{n+1})^{-1}}{ (w^{(n+1)}_{n+1})^{-1}+s^{(n)}_{t\mathrm{GP}}(x_{n+1})^2}. \label{estvt}
\end{align}
To make this implementable at step $n$, we need to remove the inaccessible $y_{n+1}$ term in both the numerator and denominator of \eqref{w}. In principle, we could attempt to (numerically) integrate the predictive distribution $Y(x_{n+1}) \sim t_{\nu}(f_t^{(n)}(x_{n+1}), \tau^2)$ against $f_t^{(n)}(x_{n+1}) \sim \mathcal{N}(\hat{f}_t^{(n)}(x_{n+1}), s_t^{(n)}(x_{n+1})^2)$; for simplicity we instead replace $(y_{n+1}-\hat{f}^{(n)}_{t\mathrm{GP}}(x_{n+1}))^2$ with its expectation: $\mathbb{E}[(y_{n+1}-\hat{f}^{(n)}_{t\mathrm{GP}}(x_{n+1}))^2] = \text{Var}[y_{n+1}] = \taun^2$ and therefore obtain the approximation
$w^{(n+1)}_{n+1} \simeq (\nu+1) \frac{ (\nu-1)\tau^2 }{ (\nu+1)^2 \tau^4} = \frac{ \nu-1}{(\nu + 1)\tau^2}$. This leads to the final look-ahead variance formula (cf.~\eqref{updategv}):
\begin{align}
\hat{s}^{(n+1)}_{t\mathrm{GP}}(x_* ; x_{n+1} )^2
&:= s^{(n)}_{t\mathrm{GP}}(x_*)^2 - \frac{v^{(n)}_{t\mathrm{GP}}(x_*,x_{n+1})^2}{(\tau^2 \frac{\nu+1}{\nu-1})+s^{(n)}_{t\mathrm{GP}}(x_{n+1})^2}.
\end{align}

\textbf{Cl-GP:} Similar to the $t$-GP, the look-ahead variance for the classification GP is intractable since  $s^{(n+1)}_{\mathrm{Cl}}$ is based on the mode $\tilde{\mathbf{\z}}^{(n+1)}_{\mathrm{Cl}}$ of the posterior $p_{\mathrm{Cl}}(\mathbf{z}|\bx,\by, x_{n+1}, y_{n+1})$. Similar to \eqref{towertmean} we use the approximation
$\tilde{\mathbf{\z}}^{(n+1)}_{\mathrm{Cl}} \simeq \check{\mathbf{z}}^{(n+1)}_{\mathrm{Cl}} :=  [\tilde{\mathbf{z}}^{(n)}_{\mathrm{Cl}}, \hat{z}_{\mathrm{Cl}}^{(n)}(x_{n+1})]$. In that case we obtain an expression similar to (\ref{estvt}), with $w^{(n+1)}_{n+1}$ replaced by $v^{(n+1)}_{n+1}$ from Eq.~(\ref{v}):
\begin{align}
\frac{{s}^{(n+1)}_{\mathrm{Cl}}(x_{n+1})^2}{s^{(n)}_{\mathrm{Cl}}(x_{n+1})^2} & \simeq \frac{(v^{(n+1)}_{n+1})^{-1}}{ (v^{(n+1)}_{n+1})^{-1}+s^{(n)}_{\mathrm{Cl}}(x_{n+1})^2}. \label{estvc}
\end{align}
The Hessian element $v^{(n+1)}_{n+1}$ is given by $$v^{(n+1)}_{n+1}= \frac{\phi( \tilde{\z}^{(n+1)}_{n+1})^2}{\Phi(\check{y}_{n+1} \tilde{\z}^{(n+1)}_{n+1})^2} +  \frac{\check{y}_{n+1} \tilde{\z}^{(n+1)}_{n+1}\phi(\tilde{\z}^{(n+1)}_{n+1})}{\Phi(\check{y}_{n+1} \tilde{\z}^{(n+1)}_{n+1})}, $$ which
depends on the next-step signed response $\check{y}_{n+1}$. To develop an approximation in terms of step-$n$ values, we once more replace $\tilde{\z}^{(n+1)}_{n+1}$ with the current mean $\hat{\z}_{\mathrm{Cl}}^{(n)}(x_{n+1})$. Moreover, the next response $\check{y}_{n+1}$ will take only two values, so $v^{(n+1)}_{n+1}$ will take on just two values $v_{n+1}^\pm$. Hence,  we can compute the ``expected value''
\begin{align}
\check{v}_{n+1} &:= v_{n+1}^+p_++v_{n+1}^-p_-,  \\
\text{where} \quad
 v_{n+1}^+ =& \frac{\phi(\hat{\z}_{\mathrm{Cl}}^{(n)}(x_{n+1}))^2}{\Phi(\hat{\z}_{\mathrm{Cl}}^{(n)}(x_{n+1}))^2} 
+ \frac{\hat{\z}_{\mathrm{Cl}}^{(n)}(x_{n+1})\phi(\hat{\z}_{\mathrm{Cl}}^{(n)}(x_{n+1}))}{\Phi(\hat{\z}_{\mathrm{Cl}}^{(n)}(x_{n+1}))}, \label{vpostrue}\\
\text{and} \quad v_{n+1}^- =& \frac{\phi(\hat{\z}_{\mathrm{Cl}}^{(n)}(x_{n+1}))^2}{\Phi(-\hat{\z}_{\mathrm{Cl}}^{(n)}(x_{n+1}))^2} 
-\frac{\hat{\z}_{\mathrm{Cl}}^{(n)}(x_{n+1})\phi(\hat{\z}_{\mathrm{Cl}}^{(n)}(x_{n+1}))}{\Phi(-\hat{\z}_{\mathrm{Cl}}^{(n)}(x_{n+1}))}, \label{vnegtrue}
\end{align}
with $p_+ := \mathbb{P}(Y(x_{n+1})> 0|\mathcal{A}_n) = \int_\mathbb{R} \Phi(\z)p_{Z(x_{n+1})}(\z|\mathcal{A}_n)d\z =  \Phi\bigg(\frac{\hat{\z}^{(n)}(x_{n+1})}{\sqrt{1+s^{(n)}_C(x_{n+1})^2}}\bigg)$, and $p_- = 1-p_+$. The final formula for the look-ahead variance becomes
$$\hat{s}^{(n+1)}_{\mathrm{Cl}}(x_{n+1})^2 :=  {s^{(n)}_{\mathrm{Cl}}(x_{n+1})^2} \cdot \frac{(\check{v}_{n+1})^{-1}}{ (\check{v}_{n+1})^{-1}+s^{(n)}_{\mathrm{Cl}}(x_{n+1})^2}.$$

\textbf{TP:} In terms of update formulas, TPs are in between GPs and $t$-GPs, with closed-form expressions available but depending on $y_{n+1}$. Specifically, the effect of adding a new observation $\left(x_{n+1}, y_{n+1} \right)$ can be highlighted in closed form, since $f(x_*)|\by, y_{n+1} \sim \mathcal{T}\left(\nu + n + 1, \hat{f}_{\mathrm{TP}}^{(n+1)}(x_*), s^{(n+1)}_{\mathrm{TP}}(x_*) \right)$, where
\begin{align}
\hat{f}_{\mathrm{TP}}^{(n+1)}(x_*) &= \hat{f}_{\mathrm{Gsn}}^{(n+1)}(x_*)\\
s^{(n+1)}_{\mathrm{TP}}(x_*)^2 &= 
\frac{\nu + \beta^{(n+1)} - 2}{\nu + n - 1} s^{(n+1)}_{\mathrm{Gsn}}(x_*)^2.
\end{align}
The effect of $y_{n+1}$ is inside $\beta^{(n+1)} = \mathbf{y}_{1:n+1}^\top [\mathbf{K}^{(n+1)}]^{-1} \mathbf{y}_{1:n+1} = \beta^{(n)} + s_{\mathrm{Gsn}}^{(n)}(\xnew)^{-1} \left(h^{(n)}(\xnew)^2  + 2 y_{n+1} h^{(n)}(\xnew) + y_{n+1}^2  \right)$ using the partition inverse equation, with $h^{(n)}(x) := -\by^\top [\mathbf{K}^{(n)}]^{-1} k(x) = -\hat{f}_{\mathrm{Gsn}}(x)$.
Since $y_{n+1}$ is unknown beforehand, we use a plugin value $\check{\beta}^{(n+1)}$ for $\beta^{(n+1)}$, relying again on the tower property:
\begin{align}
\nonumber \check{\beta}^{(n+1)}  =& \mathbb{E}[\beta^{({n+1})}|\bx, \by] \\
\nonumber =& \mathbb{E}\left[\mathbb{E}[\beta^{(n+1)} |\mathbf{x}_{1:n+1}, \mathbf{y}_{1:n+1}]|\bx, \by \right]\\
\nonumber =& \beta^{(n)} + s_{\mathrm{Gsn}}^{(n)}(\xnew)^{-2} \left(h^{(n)}(\xnew)^2 
+ 2 \hat{f}^{(n)}_{\mathrm{Gsn}}(\xnew) h^{(n)}(\xnew) +  \hat{f}^{(n)}_{\mathrm{Gsn}}(\xnew)^2 + \frac{\nu}{\nu - 2} s_{\mathrm{Gsn}}^{(n)}(\xnew)^{2} \right)\\
 = &\beta^{(n)} + \frac{\nu}{\nu - 2}.
\end{align}

\section{Gaussian Process with Monotonicity Constraint} \label{app:mgp}

Recall that since differentiation is a linear operator, the derivative  of a GP $f$ is another GP. Using $\mathbf{f}'$ as a shorthand notation for the gradient  $\nabla f$ at locations $\bx$, we have
\begin{align} \label{eq:gradient-mean}
\mathbb{E}[ \partial_{x^j} f(x_*) | \cA ] &=  \frac{\partial \mathbb{E} [f(x_*)|\cA ]}{\partial x_*^j}= \frac{ \partial \hat{f}(x_*)}{\partial x_*^j};\\
\Cov( \partial_{x^j} f(x_*), f(x_*') | \cA) &= K_{\mathbf{f}',\mathbf{f}}(x_*,x_*') = \frac{\partial}{\partial x_*^{j}} K(x_*,x_*') \\
\text{and} \quad \Cov( \partial_{x^j} f(x_*, \partial_{x^{j'}} f(x_*') | \cA ) &= K_{\mathbf{f}',\mathbf{f}'}(x_*,x_*') = \frac{\partial^2}{\partial x_*^{j} \partial (x_*')^{j'}} K(x_*,x_*').
\end{align}

In addition to the data set $(\bx, \by)$, we now introduce virtual observations $(\xv,\yv)$ with the dummy responses $y_{v,i} \in \{ -1, 1\} \times \{1,\ldots, d\}$ set according to whether $f$ is required to be decreasing ($y_{v,i} = (-1,j)$) or increasing ($y_{v,i} = (+1,j)$) with respect to the $j$th input dimension at input $x_{v,i}$. The key ``trick'' is to use a probit likelihood  $p(y_{v,i}= (+1,j) |\bx,\xv) = \Phi (\frac{1}{\eta} \partial_{x^j} f(x_{v,i}) )$,  where the small parameter $\eta$ controls the strictness of the monotonicity constraint~\citep{riihimaki2010gaussian}. The probit function approaches the Heaviside step function when $\eta \to 0$ and forces the fitted $\partial_{x^j} \hat{f}(x_{v,i})$ (computed via \eqref{eq:gradient-mean}) to match during likelihood maximization the predetermined sign of $y_{v,i}$.  An adaptive method to sequentially add the virtual inputs $\xv$ is  suggested in \cite{riihimaki2010gaussian}. Note that if there are multiple monotonic dimensions, then the same $x_{v,i}$ might be reused multiple times to satisfy the constraints on $\partial_{x^j} \hat{f}$ across different $j$-coordinates, leading to a replicated design. We also remark that monotonic metamodels are more expensive to run, since they require the use of virtual observations that increase the effective sample size to $(\bx, \xv)$ and hence require inversion of larger $\bm{K}$-matrices.

The joint prior for $\mathbf{f}$ and its gradient $\mathbf{f}'$ is given by
\begin{align}
p_{\mathrm{Mon}}\left(\left[\begin{array}{ccc}
\mathbf{f}\\
\mathbf{f}'
\end{array} \right] \big|\, \bx,\xv \right) = 
\mathcal{N}(\mathbf{0}, \mathbf{K}_{\text{joint}}), \label{2.3.1}
\end{align}
where
$
\mathbf{K}_{\text{joint}} = \left[\begin{array}{ccc}
K_{\mathbf{f},\mathbf{f}}(\bx,\bx) & K_{\mathbf{f},\mathbf{f}'}(\bx,\xv)\\
K_{\mathbf{f}',\mathbf{f}}(\xv,\bx) & K_{\mathbf{f}',\mathbf{f}'}(\xv,\xv)
\end{array}\right]$.

Using Bayes rule, the joint posterior is then %
\begin{align}
p_{\mathrm{Mon}}(\mathbf{f},\mathbf{f}'|\bx,\by,\xv,\yv) 
= \frac{p_{\mathrm{Mon}}(\mathbf{f},\mathbf{f}'|\bx,\xv)p(\by|\mathbf{f})\prod_{i} \Phi \left( y_{v,i} \partial_{x_j} f(x_{v,i}) \frac{1}{\eta}\right)}{p(\by, \yv|\bx,\xv)}. \label{2.3.2}
\end{align}

Like for the classification GP to handle the non-Gaussian terms $p({y}_{v,i}|\, \partial_{x_j} {f}(x_{v,i}))$ we approximate them with a local Gaussian likelihood 
\begin{align}
p(\yv|\mathbf{f}') \approx q(\yv|\mathbf{f}') =  
\mathcal{N}(\tilde{\bm{\mu}}_{\mathrm{Mon}}, \tilde{\bm{\Sigma}}_{\mathrm{Mon}}).
\end{align}
We use the Expectation Propagation (EP) algorithm \cite{minka2001expectation} to determine the vector of local means $\tilde{{\mu}}^i_{\mathrm{Mon}}$,  and the diagonal EP variance matrix $\tilde{\bm{\Sigma}}_{\mathrm{Mon}}$, with local variances $(\tilde{\sigma}^i_{\mathrm{Mon}})^2$. 
Details about the computation can be found in \cite{riihimaki2010gaussian}. The approximate posterior to (\ref{2.3.2}) is a product of Gaussian distributions and is simplified to
\begin{align}
\nonumber
p_{\mathrm{Mon}}(\mathbf{f},\mathbf{f}'|\bx,\xv,\by,\yv) & \approx q_{\mathrm{Mon}}(\mathbf{f},\mathbf{f}'|\bx,\xv,\by,\yv)=
\mathcal{N}(\bm{\mu}_{\text{joint}}, \mathbf{\Sigma}_{\text{joint}}).
\end{align}
The covariance matrix is $\mathbf{\Sigma}_{\text{joint}}^{-1}=\mathbf{K}_{\text{joint}}^{-1}+\tilde{\mathbf{\Sigma}}_{\text{joint}}^{-1}$, with
$\tilde{\mathbf{\Sigma}}_{\text{joint}} = \left[\begin{array}{ccc}
\sigma^2\mathbf{I} & \mathbf{0}\\
\mathbf{0} & \tilde{\mathbf{\Sigma}}_{\mathrm{Mon}}
\end{array} \right]$,
and the posterior mean is $\bm{\mu}_{\text{joint}}=\mathbf{\Sigma}_{\text{joint}} \tilde{\mathbf{\Sigma}}_{\text{joint}}^{-1} \tilde{\bm{\mu}}_{\text{joint}}$, with
$\tilde{\bm{\mu}}_{\text{joint}} = \left[\begin{array}{ccc}
\by\\
\tilde{\bm{\mu}}_{\mathrm{Mon}}
\end{array}\right]$.

The posterior mean $\hat{f}_{\mathrm{Mon}}(x_*)$ and posterior covariance $v_{\mathrm{Mon}}(x_*,x_*')$ for the M-GP metamodel are
\begin{align}
\hat{f}_{\mathrm{Mon}}^{(n)}(x_*) =&  [k(x_*),K_{\mathbf{f},\mathbf{f}'}(x_*,\xv)]\mathbf{K}_{\text{joint}}^{-1}\bm{\mu}_{\text{joint}}, \label{mone}\\
\begin{split}
v_{\mathrm{Mon}}^{(n)}(x_*,x_*') =&  K_{\mathbf{f},\mathbf{f}}(x_*,x_*')-[k(x_*),K_{\mathbf{f},\mathbf{f}'}(x_*,\xv)] \times [\mathbf{K}_{\text{joint}}+\tilde{\mathbf{\Sigma}}_{\text{joint}}]^{-1}
\left[\begin{array}{ccc}
k(x_*)\\
K_{\mathbf{f}',\mathbf{f}}(\xv,x_*)
\end{array}\right], \label{monvar}
\end{split}
\end{align}
analogously to the standard GP prediction equations (\ref{mean}) and (\ref{cov}).

\medskip

In M-GP, replacing $f$ with $z$ and again applying the EP algorithm, we reach similar expressions for posterior mean/variance as in \eqref{mone} and~\eqref{monvar}.

Similar to the $t$-GP and Cl-GP, look-ahead variance is intractable for the monotonic GP, since the EP mean $\tilde{\bm{\mu}}_{\mathrm{Mon}}$ and variance $\tilde{\bm{\Sigma}}_{\mathrm{Mon}}$ are changing as the designs are augmented. Rewriting \eqref{monvar}, we obtain
\begin{align}
\begin{split}
\tilde{v}_{\mathrm{Mon}}^{(n)}(x_*,x_*') =&  K_{\mathbf{f},\mathbf{f}}(x_*,x_*')-[K^{(n)}_{\mathbf{f},\mathbf{f}'}(x_*,\xv), k(x_*)][\tilde{\mathbf{K}}^{(n)}_{\text{joint}}+\tilde{\tilde{\mathbf{\Sigma}}}^{(n)}_{\text{joint}}]^{-1}
\left[\begin{array}{ccc}
K^{(n)}_{\mathbf{f}',\mathbf{f}}(\xv,x_*)\\
k(x_*)
\end{array}\right],
\end{split}
\end{align}
where $\ \tilde{\mathbf{K}}^{(n)}_{\text{joint}} = \left[\begin{array}{ccc}
K^{(n)}_{\mathbf{f}',\mathbf{f}'}(\xv,\xv) & K^{(n)}_{\mathbf{f}',\mathbf{f}}(\xv,\bx)\\
K^{(n)}_{\mathbf{f},\mathbf{f}'}(\bx,\xv) & K^{(n)}_{\mathbf{f},\mathbf{f}}(\bx,\bx)
\end{array}\right] $
 and $\quad \tilde{\tilde{\mathbf{\Sigma}}}^{(n)}_{\text{joint}} = \left[\begin{array}{ccc}
\tilde{\mathbf{\Sigma}}^{(n)}_{\mathrm{Mon}} & \mathbf{0}\\
\mathbf{0} & \sigma^2\mathbf{I}_{n \times n}
\end{array} \right].$

$\bm{K}^{(n)}_{\mathbf{f}',\mathbf{f}'}(\xv,\xv)$ is the step-$n$ covariance matrix for the gradient of virtual observations, and $\tilde{\mathbf{\Sigma}}^{(n)}_{\mathrm{Mon}}$ is the approximate covariance matrix for $p_{\mathrm{Mon}}(\mathbf{y}_v|\mathbf{f}')$. When calculating the one-step-ahead variance for monotonic GP, we freeze the virtual observations and their gradient, which in consequence freezes the $K^{(n)}_{\mathbf{f}',\mathbf{f}'}(\xv,\xv)$, $ K^{(n)}_{\mathbf{f}',\mathbf{f}}(\xv,\bx)$, $K^{(n)}_{\mathbf{f},\mathbf{f}'}(\bx,\xv)$, and $\tilde{\bm{\Sigma}}^{(n)}_{\mathrm{Mon}}$ matrices. Therefore, the virtual observations are treated as fixed inputs. Then, as a new observation is added, only the last row and column of the covariance matrix are updated, while the other parts remain unchanged. This approach transforms computing the look-ahead standard deviation $s^{(n+1)}_{\mathrm{Mon}}$ into the classical Gaussian observation GP as in \eqref{updatevg}. Therefore, following exactly the same procedures discussed in Section \ref{sec:update}, similar to equations \eqref{updategv} and \eqref{updatevg}, we obtain the local updated variance $s_{\mathrm{Mon}}^{(n+1)}(x_{n+1})^2$ at $x_{n+1}$, and the step-ahead variance  $s_{\mathrm{Mon}}^{(n+1)}(x_*)^2$ at any input $x_*$:
\begin{align}
\frac{s^{(n+1)}_{\mathrm{Mon}}(x_{n+1})^2}{s^{(n)}_{\mathrm{Mon}}(x_{n+1})^2} &= \frac{\taun^2}{\taun^2+s^{(n)}_{\mathrm{Mon}}(x_{n+1})^2}, \\
s^{(n+1)}_{\mathrm{Mon}}(x_*)^2
&= s^{(n)}_{\mathrm{Mon}}(x_*)^2 - \frac{v^{(n)}_{\mathrm{Mon}}(x_*,x_{n+1})^2}{\taun^2+s^{(n)}_{\mathrm{Mon}}(x_{n+1})^2}.
\end{align}

\end{document}